\documentclass[lettersize,journal]{IEEEtran}
\usepackage{amsmath,amsfonts,amsthm}
\usepackage{algorithmic}
\usepackage{mathrsfs}
\usepackage{array}
\usepackage{textcomp}
\usepackage{stfloats}
\usepackage{url}
\usepackage{verbatim}
\usepackage{graphicx}
\usepackage{cite}
\usepackage{amssymb}
\usepackage{color}
\usepackage{booktabs}
\usepackage{cite}
\usepackage{multirow}
\usepackage{float}
\usepackage{subfigure}
\usepackage{colortbl} 
\usepackage[table]{xcolor}
\usepackage{rotating}
\usepackage[ruled]{algorithm2e}
\usepackage{threeparttable}
\usepackage[
colorlinks=true,
linkcolor=blue,
anchorcolor=blue,
urlcolor=black,
citecolor=blue
]{hyperref}

\begin{document}

\title{Efficient and Effective Deep Multi-view Subspace Clustering}

\author{Yuxiu Lin, Hui Liu, ~\IEEEmembership{Member,~IEEE, } Ren Wang, Qiang Guo, and Caiming Zhang, ~\IEEEmembership{Member,~IEEE }
\thanks{This work was supported in part by the National Natural Science Foundation of China under Grant U22A2033 and 62072274, in part by the Special Funds of Taishan Scholars Project of Shandong Province under Grant tstp20221137, and in part by the Central Guidance on Local Science and Technology Development Project under Grant YDZX2022009.}
\thanks{Yuxiu Lin, Hui Liu, and Qiang Guo are with the School of Computer Science and Technology, Shandong University of Finance and Economics, Jinan 250000, China, and also with the Digital Media Technology Key Lab of Shandong Province. (e-mail: linyx$\_$0109@126.com, liuh$\_$lh@sdufe.edu.cn, and guoqiang@sdufe.edu.cn)}
\thanks{Ren Wang is with the School of Software, Shandong University, Jinan 250000, China. (e-mail: xxlifelover@gmail.com) }
\thanks{Caiming Zhang is with the School of Software, Shandong University, Jinan 250000, China, and also with the Digital Media Technology Key Lab of Shandong Province. (e-mail: czhang@sdu.edu.cn) }}

\markboth{Journal of \LaTeX\ Class Files,~Vol.~14, No.~8, August~2021}%
{Shell \MakeLowercase{\textit{et al.}}: A Sample Article Using IEEEtran.cls for IEEE Journals}


\maketitle

\begin{abstract}
Recent multi-view subspace clustering achieves impressive results utilizing deep networks, where the self-expressive correlation is typically modeled by a fully connected (FC) layer. However, they still suffer from two limitations. i) The parameter scale of the FC layer is quadratic to sample numbers, resulting in high time and memory costs that significantly degrade their feasibility in large-scale datasets. ii) It is under-explored to extract a unified representation that simultaneously satisfies minimal sufficiency and discriminability. To this end, we propose a novel deep framework, termed Efficient and Effective deep Multi-View Subspace Clustering (E$^2$MVSC). Instead of a parameterized FC layer, we design a Relation-Metric Net that decouples network parameter scale from sample numbers for greater computational efficiency. Most importantly, the proposed method devises a multi-type auto-encoder to explicitly decouple consistent, complementary, and superfluous information from every view, which is supervised by a soft clustering assignment similarity constraint. Following information bottleneck theory and the maximal coding rate reduction principle, a sufficient yet minimal unified representation can be obtained, as well as pursuing intra-cluster aggregation and inter-cluster separability within it. Extensive experiments show that E$^2$MVSC yields comparable results to existing methods and achieves state-of-the-art performance in various types of multi-view datasets.

\end{abstract}

\begin{IEEEkeywords}
Multi-view data, representation learning, self-expression, mutual information, subspace clustering.
\end{IEEEkeywords}

\section{Introduction}
\IEEEPARstart{S}{ubspace} clustering (SC), which allocates unlabeled samples lying in a mixture of low-dimensional subspaces to the corresponding clusters, has been a crucial research topic in image processing, data mining, and machine learning communities \cite{Wang2021FastPM, Peng2018StructuredAF}. State-of-the-art SC works are mainly based upon the ``self-expressiveness (SE)'' property of subspaces, where each sample could be expressed as a linear combination of other samples inside the same subspace\cite{Ji2017DeepSC}. 

However, with the diversification of data acquisition devices, a mass of data actually takes on a multi-modal/multi-feature character in real world\cite{Xie2020JointDM, Wang2023BiNuclearTS, DBLP:journals/tip/ChenMPZP23, ZhangFHCXTX20, DBLP:journals/isci/YuLLLS23}. For example, an image could be expressed in various feature descriptors, including Local Binary Patterns (LBP), Scale Invariant Feature Transform (SIFT), and Histogram of Oriented Gradient (HOG); a webpage could be represented by images, texts, URLs, and videos. Clearly, multi-view data contain more comprehensive information as different views are complementary, yet each view has unique statistical properties, raising challenges to traditional single-view-oriented methods\cite{SunDL21}. Therefore, multi-view subspace clustering (MVSC) is derived, aiming to exploit and integrate consistent and complementary (diverse) information across views to boost performance\cite{Cao2015ConstrainedMV, Li2022HighOrderCP, Tang2022ConstrainedTR}. 

Among multitudinous MVSC methods, SE-based still performs well\cite{Zhou2020DualSM}. Earlier methods \cite{Gao2015MultiviewSC, LuoZZC18, Zheng2019FeatureCM} used shallow or linear functions on original samples to learn a shared self-expressive matrix, where the nonzero entries represent two samples that belong to the same subspace. Since noise or gross corruption is inevitable in practice, computing subspace representation directly from raw data will degrade clustering accuracy. Thus, subsequent studies \cite{Zhang2017LatentMS, Chen2020MultiViewCI} attempt to seek a low-dimensional embedding space for primitive views. However, limited by poor representability, these linear transformations are arduous to model complex nonlinear properties of high-dimensional, heterogeneous multi-view data\cite{ZhangFHCXTX20}.

Deep multi-view subspace clustering (DMVSC) bridges this gap by leveraging neural networks, specifically auto-encoders (AEs), for feature extraction and representation. A typical DMVSC framework is that\cite{Abavisani2018DeepMS}, first pre-training a set of view-specific AEs to independently learn latent representations of different views. Then, during the fine-tuning stage, an FC layer without a bias is embedded into AEs to simulate the self-expression process. The trainable FC layer parameters correspond to the self-expressive coefficients. Following this pipeline, substantial variants are proposed for better mining and fusing multi-view information, such as introducing extra regularization terms \cite{Li2019ReciprocalMS, Wang2021DeepMS} or designing various feature fusion strategies\cite{LuLZ21, Wang2022SelfSupervisedIB}. Despite promising advancements, an unavoidable limitation is that the number of trainable parameters in the FC layer scales quadratically with sample size $n$. This could incur an expensive computational cost, especially for considerably large-scale data\cite{DBLP:journals/tcyb/LiLZLF22}.

A few studies have proposed the anchor-based strategy to explore fast clustering for large-scale data \cite{KangZZSHX20, Sun2021ScalableMS, Wang2021FastPM}. By selecting a set of representative anchor points to represent the entire data, such approaches reduce the time cost to linear complexity. However, it is challenging to determine the selection mechanism and numbers of anchor samples\cite{Li2020MultiviewCA}, which inevitably sacrifices some clustering accuracy. Besides, to the best of our knowledge, in SE-based DMVSC, almost no research focuses on extending the applicability of the conventional self-expression model to large-scale scenarios.

\begin{figure*}[!t]
\centering
\includegraphics[width=\linewidth]{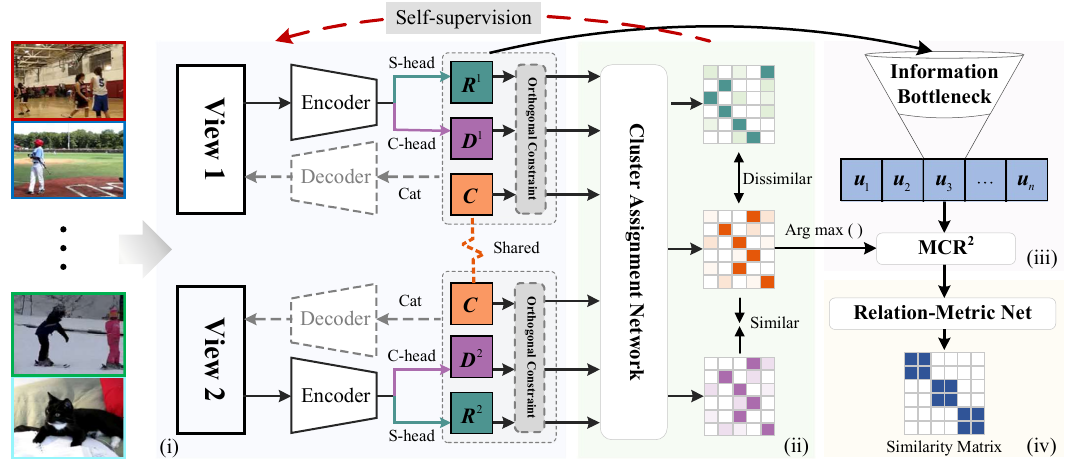}
\caption{Overview of the proposed E$^2$MVSC, two views of input images as an example. Four modules are mainly concerned. (i) Multi-type latent representation learning first extracts latent representations for each view using individual auto-encoders. The encoder has two heads (S-head and C-head) for generating the view-superfluous $\boldsymbol{R}^{v}$ and view-complementary representation $\boldsymbol{D}^{v}$, respectively. (ii) Soft cluster assignment supervision targets obtaining the soft clustering assignment to supervise the feature decoupling learning. (iii) Unified representation learning constructs a minimal sufficiency unified representation $\boldsymbol{U}=\{\boldsymbol{u}_{i}\}_{i=1}^{n}$ based on the information bottleneck, and further boosts discrimination with the MCR$^2$ principle. (iv) Self-expression learning employs the Relation-Metric Net to efficiently obtain similarity scores of any sample pairs as corresponding self-expressive coefficients. Note that Cat is concatenation for short.} 
\label{fig1}
\end{figure*}

Beyond efficiency, the clustering effectiveness of DMVSC also has room for improvement. It is believed that integrating latent representations of all views will lead to a more reliable feature representation\cite{Zhang2019AE2NetsAI, Zheng2022ComprehensiveMR}. Existing methods design various feature fusion processes \cite{Wang2020iCmSCIC, LuLZ21} or enforce latent representations to be consistent \cite{XuT0P0022} to embed a unified feature space, but most of them cannot guarantee that consistency and complementarity principles are completely preserved. In fact, a unified representation that is clustering-oriented should possess two properties, i.e., minimal sufficiency \cite{Lin2023DualCP} and discriminability \cite{Cai2022UnsupervisedDD, Fu2023MultiViewCF}. Concretely, minimal-sufficient representation refers to fully exploring the consistent and complementary information from multi-view data, while reducing task-irrelevant, superfluous information (e.g., background or texture pixels in an image). Further, the discriminatory representation needs to satisfy intra-cluster discrepancy minimization as well as inter-cluster margin maximization, for partitioning data points around cluster boundaries more precisely. However, relatively little work takes all the properties mentioned into consideration, limiting their representability.
 
Based on the above concerns, this paper presents a novel DMVSC method, namely E$^2$MVSC, to enhance \textbf{efficiency} (w.r.t. self-expression learning) and \textbf{effectiveness} (w.r.t. unified representation learning) for clustering multi-view data. Fig. \ref{fig1} illustrates the overall flowchart of E$^2$MVSC. Specifically, we first design a multi-type latent representation learning module and a soft clustering assignment supervision module, explicitly decoupling consistent, complementary, and superfluous features from low-dimensional embedding space for each view. Then, drawing inspiration from information theory (via information bottleneck (IB)\cite{DBLP:conf/iclr/Federici0FKA20} and maximal coding rate reduction (MCR$^2$) \cite{Yu2020LearningDA}), we strive for a reliable unified representation that well models consistency, complementarity, and discrimination. Finally, we develop a deep metric network termed Relation-Metric Net to directly and efficiently measure the similarity relation between any sample pairs, which casts off the generic self-expression learning layer with high computational complexity in existing methods.

The contributions of this work are summarized as follows:

\begin{itemize}
    \item We explicitly decouple consistent, complementary, and superfluous information across multi-view data in a self-supervised manner, to properly infer a minimal-sufficient unified representation. 
    
    \item We further introduce the MCR$^2$ principle to enhance both intra-cluster aggregation and inter-cluster separability of unified representation, which is necessary for clustering effectiveness.
   
    \item We replace the FC layer in conventional self-expression models with a deep metric network, which disentangles the quadratic correlation between parameter scale and sample numbers, promoting model efficiency.
    \item Extensive experiments show that E$^2$MVSC outperforms its competitors across seven benchmark datasets, especially enjoying superior performance in large-scale scenarios.

\end{itemize}

The rest of this paper is organized as follows. Section \ref{Related Work} introduces notations and reviews a few important works most related to our proposed method. Detailed analysis of the proposed E$^2$MVSC is provided in Section \ref{Method}. Sufficient experiments conducted on seven benchmark datasets are reported in Section \ref{Experiments}. At last, the conclusions and future work are presented in Section \ref{Conclusion}.

\section{Notations and Related Work}
\label{Related Work}
\subsection{Notations and Preliminaries}
In this paper, we denote all the matrices with capital boldfaced letters (e.g., $\boldsymbol{A}$), vectors with lower-case boldfaced letters (e.g., $\boldsymbol{a}$), scalars with italic lower-case letters(e.g., $a$), and sets with calligraphic letters (e.g., $\mathcal{A}$), respectively. For clarity, the involved mathematical notations and explanations used throughout this paper are summarized in Table \ref{tab1}. 

Suppose that we have multi-view observations from $V$ views, $\mathcal{X}=\left\{{\boldsymbol{X}^{1}}, {\boldsymbol{X}^{2}},\cdots , {\boldsymbol{X}^{V}}\right\}$, where $\boldsymbol{X}^{v}\in \mathbb{R}^{d_v \times n}$ denotes the data matrix corresponding to $v$-th view with $n$ $d_v$-dimensional samples, $\boldsymbol{x}_{i}^{v}\in {{\mathbb{R}}^{{{d}_{v}}}}$ is the $i$-th instance of $\boldsymbol{X}^{v}$. SE-based multi-view subspace clustering aims to obtain a comprehensive self-expressive coefficient matrix (also dubbed subspace representation) $\boldsymbol{S}\in {{\mathbb{R}}^{{n \times n}}}$ shared by all views and, from this, accurately partition the samples into $K$ clusters.

\begin{table}[!t]
\centering
\caption{Main notations throughout the paper}
\label{tab1}
\footnotesize
\tabcolsep 10.5pt 
\begin{tabular}{ll} 
\toprule
\textbf{Notation} & \textbf{Definition}                            \\ 
\midrule
$d_v$     & Dimension of the feature vector in view $v$.   \\
$d$     & Dimension of the latent representation.   \\
$\boldsymbol{I}$     & Identity matrix.     \\ 
$\boldsymbol{1}$     & A vector with all ones.     \\
$\text{diag}(\cdot)$               & Vector of the diagonal elements of a matrix. \\ 
$\left| \cdot \right|$   & Absolute operator.\\
${I}\left( \cdot; \cdot \right)$     & The mutual information between two variables. \\
$p(\cdot)$  & The distribution of certain variable. \\ 
$tr(\cdot)$  & Trace operation. \\ 
\bottomrule
\end{tabular}
\end{table}

\subsection{Deep Multi-view Subspace Clustering}
With excellent representation ability, deep multi-view clustering methods are ubiquitously studied in various communities, such as image processing, machine learning, data mining, and so on. The representative works, including DMSC\cite{Abavisani2018DeepMS}, MvSCN\cite{Huang2019MultiviewSC}, RMSL\cite{Li2019ReciprocalMS}, gLMSC\cite{ZhangFHCXTX20}, DMSC-UDL\cite{Wang2021DeepMS}, SDMSC\cite{DBLP:journals/tcyb/LiLZLF22}, SIB-MSC\cite{Wang2022SelfSupervisedIB}, SDMVC\cite{Xu2023SelfSupervisedDF}, etc. Several are briefly introduced here. RMSL\cite{Li2019ReciprocalMS} exploits hierarchical self-representative layers to learn view-specific subspace representations, which are encoded into a common representation through backward encoding networks. SDMVC\cite{Xu2023SelfSupervisedDF} concatenates multi-view features as global features and designs a self-supervised mechanism to guide discriminative feature learning. Among these approaches, SE-based DMVSC is fairly effective for high-dimensional data and shows great promise. Without losing generality, they usually apply view-specific AEs to build an end-to-end latent feature extraction framework and integrate multi-view information into one unified representation. Then, a self-expression learning layer (an FC layer without a bias) is inserted into AEs to learn a consistent subspace representation. Regardless of the unified representation learning process, the classical SE-based DMVSC framework can be defined as the following optimization problem:
\begin{equation}
 \label{eq1}
   \begin{split}
    &\underset{\boldsymbol{S}}{\mathop{\min }}\,\sum\nolimits_{v=1}^{V}{\left\| {\boldsymbol{{X}}^{v}}-\hat{\boldsymbol{{X}}^{v}} \right\|_{F}^{2}} \text{      } + \lambda_1{\left\| {\boldsymbol{{U}}}-{\boldsymbol{{U}}}\boldsymbol{S} \right\|_{F}^{2}}\\
    &+ \lambda_2 R(\boldsymbol{S})   \qquad
 \mathrm{s.t.}\text{      }diag(\boldsymbol{S})=0, \boldsymbol{S}^T\boldsymbol{1}=\boldsymbol{1},
   \end{split}
\end{equation}
where ${{\left\| \cdot  \right\|}_{F}}$ denotes the Frobenius norm. ${\boldsymbol{U}}\in {{\mathbb{R}}^{{d \times n}}}$ and $\hat{\boldsymbol{X}}^{v}\in \mathbb{R}^{d_v  \times n}$ are the unified representation integrated from multiple views and the reconstruction output corresponding to view $v$, respectively. $R( \cdot  )$ is a certain regularization term, e.g., nuclear norm, ${\ell }_1$ norm, to achieve an ideal solution. $\lambda_1,\lambda_2>0$ are trade-off parameters. Additional constraint $diag(\boldsymbol{S})=0$ is used to prevent the trivial solution $\boldsymbol{S}=\boldsymbol{I}$, that is, to exclude the case where each sample is represented by itself. The other constraint $\boldsymbol{S}^T\boldsymbol{1}=\boldsymbol{1}$ ensures the normalization of $\boldsymbol{S}$, indicating that samples lie in a union of affine subspaces, rather than linear subspaces\cite{Gao2015MultiviewSC}. Once $\boldsymbol{S}$ is determined, the affinity matrix $\boldsymbol{A}\in {{\mathbb{R}}^{{n \times n}}} =\left| \boldsymbol{S} \right|+| {\boldsymbol{S}^{T}} | $ is computed as input to spectral clustering\cite{Abavisani2018DeepMS}. 

\begin{figure}[h]
\centering
\includegraphics[width=\linewidth]{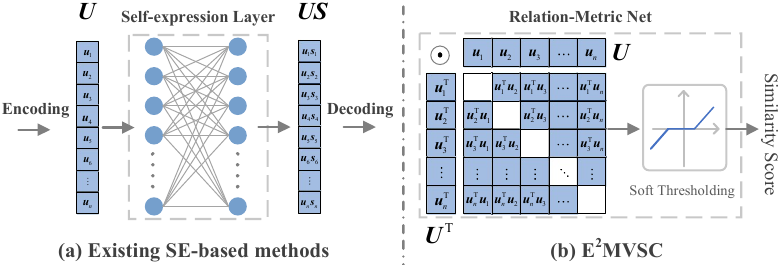}
\caption{Schematic diagram of self-expressive matrix learning models.}
\label{fig2}
\end{figure}

In the above framework, the solution of $\boldsymbol{S}$ is parameterized to the weights of a self-expression learning layer, as shown in Fig. \ref{fig2}(a). This requires putting the entire sample set into one batch for training at a time. Hence, the coefficient matrix available in memory has a quadratic size ($n\times n$) with sample numbers, which is computationally prohibitive and not specific to large-scale data clustering scenarios. 

\subsection{Large-scale Multi-view Subspace Clustering}

To alleviate high time and memory costs, multifarious large-scale MVC methods built upon matrix factorization \cite{Liu2021OnepassMC, Wan2023AutoweightedMC} or anchor graph strategies \cite{KangZZSHX20, Li2020MultiviewCA, Wang2021FastPM} are proposed. The anchor selection strategy has proven to be a useful and suitable option for handling large-scale datasets\cite{Zhang2023CenterCG}. It samples a set of representative points as anchors to approximate the entire data distribution, and then performs self-expression for anchors. LMVSC\cite{KangZZSHX20} first extends the anchor-based approach to MVSC scenarios, compressing the training of the model to linear time. Subsequently, SMVSC\cite{Sun2021ScalableMS} jointly integrates anchor learning and subspace learning into a unified framework, resulting in linear order complexity for large-scale data. FPMVS-CAG\cite{Wang2021FastPM} offers a parameter-free multi-view subspace clustering framework with the aid of consensus anchors and scales well for large datasets. EOMSC-CA\cite{Liu2022EfficientOM} proposed a unified framework with optimizing anchor selection and subspace graph construction jointly.

It should be noted that, while anchor-based variants can lead to time savings, selected anchors through heuristic sampling strategies may be of low quality, finally reflecting in erroneous bipartite graph structures. Recent research (SSCN \cite{Busch2020LearningSM} and SENet \cite{ZhangYVL21}) decouples the parameter scale from sample numbers by treating self-expressive coefficients as a network output instead of trainable network parameters. Although such methods show promising performance, they are single-view oriented. Due to the complexity of heterogeneous view features, it fails to go for multi-view data directly. Moreover, in SE-based DMVSC methods, there is a dearth of research on extending self-expression models to large-scale datasets.  

\subsection{Multi-view Representation Learning}
Multi-view representation learning \cite{Mao2021DeepMI, Wan2021MultiViewIR, Wang_2023_CVPR} aims to integrate information across views to comprehensively describe objects, which is the bedrock for most MVSC tasks as clustering performance profits from it. AE$^2$-Nets\cite{Zhang2019AE2NetsAI} integrates intrinsic information from heterogeneous views into an intact latent representation for clustering, utilizing a nested auto-encoder framework. gLMSC\cite{ZhangFHCXTX20} assumes that different views can be reconstructed based on a common representation. Since excavating too much information, including superfluous ones, finding cluster structure in latent space is inappropriate. MFLVC \cite{XuT0P0022} proposes a multi-level feature learning framework, implicitly filters out view-private information during the network learning process, and gains common semantics by imposing contrastive learning on high-level features. Contrastive learning is a hot topic in multi-view communities, but it treats the same sample across different views as positive pairs that are too restricted, i.e., maximizing the consistency of any two representations while losing sight of diversity.

In addition to contrastive learning, another alternative strategy is based on information theory. MIB \cite{DBLP:conf/iclr/Federici0FKA20} and DMIM \cite{Mao2021DeepMI} identify shared and superfluous information from diverse views, incorporating information bottleneck theory, to obtain a robust representation. Similarly, CMIB-Nets \cite{Wan2021MultiViewIR} and MSCIB \cite{Yan2023MultiviewSC} explore the shared latent structure and intra-view intrinsic information with the help of information bottleneck theory. However, the above methods assume information other than consistent semantic information is superfluous, again ignoring complementary information across views and thus compromising clustering effects. In addition, the research on discriminability (summarized as similarity contracts and dissimilarity contrasts) in unified representation is insufficient. IMVC \cite{Fu2023MultiViewCF} achieves a unified feature representation that is discriminative by maximizing mutual information between instances and their \textit{k}-nearest neighbors. 

\section{The Proposed Method} 
\label{Method}
\subsection{Motivation}
As mentioned above, there are still two limitations that exiting DMVSC methods fail to work out simultaneously.

\subsubsection{Effective Feature Representation Learning}Clustering effects are sensitive to the goodness of learned feature representation\cite{Ren2022DeepCA}. Precisely, a desired unified representation is expected to satisfy the following two compatible properties, which were not well satisfied in previous methods. \textbf{\textit{Minimal-Sufficiency}}: a competent feature representation ought to possess sufficiency and compactness. In which, as much useful information of the clustering-relevant (consistency and complementarity) as possible should be preserved, while the irrelevant view information (redundancy) is filtered out. \textbf{\textit{Discriminability}}: intra-cluster aggregation and inter-cluster separability. As the name implies, feature representations of samples from the same cluster should be highly similar since they, in a sense, belong to a low-dimensional linear subspace, while feature representations of samples among different clusters should be maximally dissimilar and separable from each other.

\subsubsection{Efficient Self-expression Learning} Most existing SE-based DMVSC methods severely suffer from high time costs and memory footprints, mainly because the FC layer requires all samples to be loaded into one batch for learning self-expressive coefficients. Especially in large-scale datasets, it is computationally infeasible. In addition, we find that few studies are conducted for the self-expression learning model in deep multi-view subspace clustering, consequently limiting its applicability in realistic large-scale data scenarios. Therefore, developing DMVSC algorithms that are scalable and efficient is necessary.

\subsection{A Base Assumption}
Our method follows a basic assumption within multi-view data: each view can be viewed as encompassing three distinct information components that are orthogonal to each other, as shown in Fig. \ref{fig3}(a). One is view-consistent information that signifies a consensus of knowledge across all views. Another two are private for each view. View-complementary information, denoting some knowledge in one view, facilitates describing data comprehensively, yet other views do not possess it. View-superfluous information refers to irrelevant and even misleading knowledge for clustering, e.g., background pixels in an image. Based on this assumption, we intend to decouple all information and discard view-superfluous information to generate a minimal-sufficient unified representation for self-expressive coefficient learning.

\subsection{The Proposed E$^2$MVSC}
\label{subsectionC}
Given multi-view data, the proposed method explores how to achieve effective unified representation learning and efficient self-expressive coefficient learning. As in Figure \ref{fig1}, E$^2$MVSC mainly consists of four primary modules. All modules will be illustrated one by one.

\subsubsection{Multi-type Latent Representation Learning Module} 
Raw multi-view data space contains adverse information such as random noise, and feature dimensions are different in heterogeneous views. Hence, extracting latent features of the same dimension out of it, ${{\mathbb{R}}^{{{d}_{v}}}}\to {{\mathbb{R}}^{d}}$, is important for multi-view clustering. $d$ is the dimension of latent representations. Most existing works, like\cite{Abavisani2018DeepMS, Wang2021DeepMS, Fu2023MultiViewCF}, normally utilize view-specific AEs and optimize a reconstruction loss to learn an intact latent representation consistent with the raw data. However, the low-dimensional latent representation, learned in this form, almost mixes the entire information about the original view, including redundancy. Therefore, we here seek to decompose three kinds of information components from latent space for each view.

To this end, we first design a multi-type latent representation learning module, which consists of multiple view-specific encoders and decoders. Two heads are added on top of each encoder output to extract two different latent representations, respectively. A view-complementary representation learning head (C-head for short) encodes complementary information, and a view-superfluous representation learning head (S-head for short) encodes redundant information. Taking the $v$-th view as an example, with $\boldsymbol{X}^{v}$ as input, the view-superfluous representation $\boldsymbol{R}^{v}\in \mathbb{R}^{d \times n}$ and the view-complementary representation $\boldsymbol{D}^{v}\in \mathbb{R}^{d \times n}$ are given by: 
\begin{equation}
\label{eq3}
    \begin{split}
        &\boldsymbol{R}^{v}:= {S_{v}}\left( \boldsymbol{X}^{v}; {\boldsymbol{\Theta} _{e,b}^{v}}, {\boldsymbol{\Theta} _{e,s}^{v}} \right)\\
        &\boldsymbol{D}^{v}:= {C_{v}}\left( \boldsymbol{X}^{v}; {\boldsymbol{\Theta} _{e,b}^{v}}, {\boldsymbol{\Theta} _{e,c}^{v}} \right), 
    \end{split}
\end{equation}
where $S_{v}\left(\right)$ and $C_{v}()$, refer to the mapping functions of S-head and C-head. ${\boldsymbol{\Theta} _{e,b}^{v}}$, ${\boldsymbol{\Theta} _{e,s}^{v}}$, and ${\boldsymbol{\Theta} _{e,c}^{v}}$ denote the parameter sets of the encoder backbone network, S-head, and C-head, respectively. For convenience, we introduce ${\boldsymbol{\Theta} _{e}^{v}}=\left\{ \boldsymbol{\Theta} _{e,b}^{v},\boldsymbol{\Theta} _{e,s}^{v},\boldsymbol{\Theta} _{e,c}^{v} \right\}$ to represent all trainable parameters in the encoder. 

Since a view-consistent representation $\boldsymbol{C}\in \mathbb{R}^{d \times n}$ reveals underlying knowledge shared by all views, it is reasonable that we regenerate all raw views from $\boldsymbol{C}$. According to the degradation learning strategy \cite{Wan2021MultiViewIR, Zhou2023SemanticallyCM}, $\boldsymbol{C}$ is defined as a set of trainable parameters and starts with random initialization. Here, we model an intact latent embedding by concatenating three representations, i.e., ${\boldsymbol{Z}^{v}={\left[\boldsymbol{C};\boldsymbol{D}^{v};\boldsymbol{R}^{v}\right]}}\in \mathbb{R}^{3d \times n}$, as decoder input. The decoding process is formulated as ${D_{v}}\left( \boldsymbol{Z}^{v}; {\boldsymbol{\Theta}_{d}^{v}} \right): {{\mathbb{R}}^{3d}}\to {{\mathbb{R}}^{{{d}_{v}}}}$ with learnable parameters ${\boldsymbol{\Theta}_{d}^{v}}$.

To optimize networks, we can minimize errors between raw views $\boldsymbol{X}^{v}$ and corresponding decoded views $\hat{\boldsymbol{X}}^{v}\in \mathbb{R}^{d_v  \times n}$. Thus, the within-view reconstruction loss in latent space will be given to the objective function, with the form
\begin{equation}
\label{eq4}
    \begin{split}
        &{\mathcal{L}_{AEs}}:=\sum\nolimits_{v=1}^{V}{\left\| {\boldsymbol{X}^{v}}-{{{\hat{\boldsymbol{X}}}}^{v}} \right\|_{F}^{2}} \\
        &+ \frac{1}{N}\sum\nolimits_{v=1}^{V}{\left( {{\left\langle \boldsymbol{C}, \boldsymbol{D}^{v} \right\rangle }^{2}} +{{\left\langle \boldsymbol{C}, \boldsymbol{R}^{v} \right\rangle }^{2}} + {{\left\langle \boldsymbol{D}^{v},\boldsymbol{R}^{v} \right\rangle }^{2}} \right)}, 
    \end{split}
\end{equation}
where the second term, minimizing the inner product between any pairs of representations, i.e., ${\left\langle \cdot,\cdot \right\rangle} \to {0}$, is used to constrain latent representations to be orthogonal to each other. We symbolize the second term as $\mathcal{L}_{ortho}$, for the convenience of conducting ablation studies in experiments.

\subsubsection{Soft Cluster Assignment Supervision Module}
Although we have captured three sets of feature representations for every view by Eq.(\ref{eq4}), two view-private components (redundancy and complementary) cannot be clearly distinguished in the process. This is due to the absence of labels in unsupervised settings, which highlights the difficulty of identifying which information is task-irrelevant. Therefore, additional constraints need to be incorporated into the representation learning process to regularize these two components.

\begin{figure}[h]
\centering
\includegraphics[width=\linewidth]{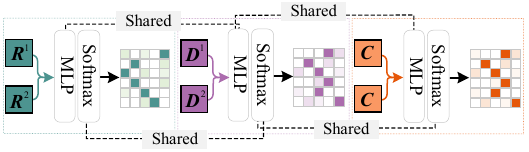}
\caption{The cluster assignment network in our proposed E$^2$MVSC by taking the example of two views.}
\label{fig2_1}
\end{figure}

Inspired by self-supervised learning, we further propose a soft cluster assignment supervision module that guides latent representation learning to disentangle two private components. Following \cite{Mao2021DeepMI, XuT0P0022}, a cluster assignment network with a two-layer multi-layer perceptron (MLP) is introduced. Notably, the cluster assignment network, which acts on latent space, is shared across different representations, aiming to guarantee that all representations follow the same projection rule. Taking the view-consistent representation as input of MLP, and stacking the softmax operator at the last layer, the consistent probability distribution matrix $\boldsymbol{Q}\in \mathbb{R}^{n \times K}$ of cluster assignment will be calculated:
\begin{equation}
\label{eq5}
    \begin{split}
        &\boldsymbol{Q}:= {MLP}\left( \boldsymbol{C}; {\boldsymbol{\Theta}_{clu}}\right), 
    \end{split}
\end{equation}
where ${\boldsymbol{\Theta}_{clu}}$ are learnable parameters of the cluster assignment network. Likewise, we concatenate all view-superfluous representations $\boldsymbol{R}^{v}$ and all view-complementary representations $\boldsymbol{D}^{v}$, to form a global superfluous representation and a global complementary representation, respectively. Then, feeding them into MLP to calculate corresponding cluster assignment probability distribution matrices, $\boldsymbol{Q}^{R}\in \mathbb{R}^{n \times K}$ and $\boldsymbol{Q}^{D}\in \mathbb{R}^{n \times K}$, as shown in Fig. \ref{fig2_1}. Each entry $\boldsymbol{q}^{R}_{i,k}$, $\boldsymbol{q}^{D}_{i,k}$, and $\boldsymbol{q}_{i,k}$ signals the probability that allocates the $i$-th sample to the $k$-th cluster via the corresponding representation. 

In principle, cluster assignments derived from consistent or complementary representations should be similar, but dissimilar from superfluous representations. With this end in view, a soft cluster assignment similarity constraint is imposed here, serving as a self-supervised signal to refine the differential learning of complementary and superfluous representations. Specifically, we introduce the mean square error to measure the divergence between any two cluster assignment matrices. The similarity constraint loss is defined as:
\begin{equation}
\label{eq6}
    \begin{split}
        {\mathcal{L}_{SS}}:={\left\| {\boldsymbol{Q}}-{\boldsymbol{Q}^{D}} \right\|_{F}^{2}}- {\left\| {\boldsymbol{Q}}-{\boldsymbol{Q}^{R}} \right\|_{F}^{2}}.
    \end{split}
\end{equation}

\subsubsection{Unified Representation Learning Module}
\begin{figure}[!t]
\centering
\includegraphics[width=\linewidth]{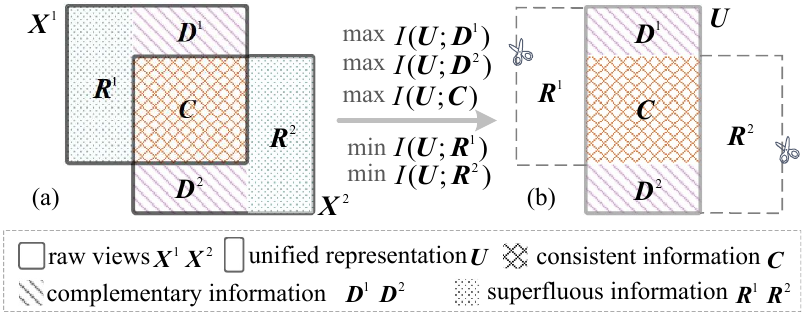}
\caption{Visualization of the multi-view information compression process. Here we consider the case of two views for clarification. (a) All information contained in raw views $\boldsymbol{X}^{1}\cup{\boldsymbol{X}^{2}}$, including clustering-relevant information and superfluous information. (b) The minimal-sufficient unified representation $\boldsymbol{U}$ (grey rectangle box) is finally achieved by information bottleneck theory, which integrates consistent and diverse information embedded in individual views, as well as mitigating the influence of superfluous information. }
\label{fig3}
\end{figure}

After multi-type latent representations are successfully extracted, we are inspired by information bottleneck theory to model a minimal yet most informative unified representation, $\boldsymbol{U}\in \mathbb{R}^{d \times n}$, across all views. Concretely, the unified representation is encouraged to preserve clustering-relevant information as much as possible, i.e., maximize mutual information ${I}\left({\boldsymbol{U}}; \boldsymbol{C} \right)$ and ${I}\left({\boldsymbol{U}}; \boldsymbol{D}^{v} \right)$. This will enable information sufficiency. Meanwhile, the irrelevant information should be reduced from $\boldsymbol{U}$ to maintain minimality, i.e., minimize mutual information ${I}\left({\boldsymbol{U}}; \boldsymbol{R}^{v} \right)$. As shown in Fig. \ref{fig3}, the information bottleneck loss can be written as the following formula:
\begin{equation}
\label{eq7}
    \begin{split}
        {\mathcal{L}_{IB}}:=\sum\nolimits_{v=1}^{V}({{I}\left({\boldsymbol{U}}; \boldsymbol{R}^{v} \right)-{I}\left({\boldsymbol{U}}; \boldsymbol{D}^{v} \right)})-{I}\left({\boldsymbol{U}}; \boldsymbol{C} \right).\\
    \end{split}
\end{equation}

Now, we present the concrete optimization of Eq.(\ref{eq7}) in detail. Start with ${I}\left( \boldsymbol{U}; {\boldsymbol{R}^{v}} \right)$ minimization, which facilitates filtering out superfluous information from the $v$-th view. According to the definition of mutual information, we have
\begin{equation}
\label{eq8}
    \begin{split}
          {I}\left( \boldsymbol{U}; {\boldsymbol{R}^{v}} \right)& =\iint{p({\boldsymbol{u}},\boldsymbol{r}^{v})}log\frac{p({\boldsymbol{u}},\boldsymbol{r}^{v})} {p({\boldsymbol{u}})p(\boldsymbol{r}^{v})}d\boldsymbol{u}d\boldsymbol{r}^{v} \\ 
          & =\iint{p({\boldsymbol{u}},\boldsymbol{r}^{v})}log\frac{p({\boldsymbol{u}}|\boldsymbol{r}^{v})}{p(\boldsymbol{u})}d\boldsymbol{u}d\boldsymbol{r}^{v} , \\ 
    \end{split}
\end{equation}
where $\boldsymbol{u}$ and ${\boldsymbol{r}^{v}}$ referred to an arbitrary element in the unified representation $\boldsymbol{U}$ and the view-superfluous representation $\boldsymbol{R}^{v}$, respectively. Given the intractability of directly calculating ${I}\left( \boldsymbol{U}; {\boldsymbol{R}^{v}} \right)$, we here follow the variational strategy\cite{Barber2003TheIA} that optimizes the variational upper bound of mutual information, aiming to acquire an approximate solution as an alternative. To be specific, we introduce $q(\boldsymbol{u})$ as a variational estimation of the marginal distribution $p(\boldsymbol{u})$ for convenient computing. In view of the non-negativeness of Kullback-Leibler (KL) divergence, it naturally leads to 
\begin{equation}
\label{eq9}
    \begin{split}
         & KL\left[ p(\boldsymbol{u})||q(\boldsymbol{u}) \right]\ge 0\Rightarrow \int{p(\boldsymbol{u})\log \frac{p(\boldsymbol{u})}{q(\boldsymbol{u})}}d\boldsymbol{u}\ge 0 \\ 
         & \Rightarrow \int{p(\boldsymbol{u})\log p(\boldsymbol{u})}d\boldsymbol{u}\ge \int{p(\boldsymbol{u})\log q(\boldsymbol{u})}d\boldsymbol{u}. \\  
    \end{split}
\end{equation}

Using the inequality of Eq.(\ref{eq9}), Eq.(\ref{eq8}) is deduced with a variation upper bound as follows:
\begin{equation}
\label{eq10}
    \begin{split}
         {I}\left( \boldsymbol{U}; {\boldsymbol{R}^{v}} \right)& =\iint{p({\boldsymbol{u}},\boldsymbol{r}^{v})}log\frac{p({\boldsymbol{u}}|\boldsymbol{r}^{v})}{p(\boldsymbol{u})}d\boldsymbol{u} d\boldsymbol{r}^{v} \\ 
         & \le \iint{p({\boldsymbol{u}},\boldsymbol{r}^{v})}log\frac{p({\boldsymbol{u}}|\boldsymbol{r}^{v})}{q(\boldsymbol{u})}d\boldsymbol{u}d\boldsymbol{r}^{v}  \\ 
         & =\int{d\boldsymbol{r}^{v}}p({\boldsymbol{r}^{v}})\int{d{\boldsymbol{u}}p({\boldsymbol{u}}|\boldsymbol{r}^{v})log\frac{p({\boldsymbol{u}}|\boldsymbol{r}^{v})}{q(\boldsymbol{u})}}. \\ 
    \end{split}
\end{equation}

Next, we consider calculating the variation lower bound of ${I}\left( \boldsymbol{U}; {\boldsymbol{D}^{v}} \right)$. Based on Eq.(\ref{eq8}), we define
\begin{equation}
\label{eq11}
    \begin{split}
          {I}\left( \boldsymbol{U}; {\boldsymbol{D}^{v}} \right)=\iint{p({\boldsymbol{u}},\boldsymbol{d}^{v})}log\frac{p({\boldsymbol{u}}|\boldsymbol{d}^{v})}{p(\boldsymbol{u})}d\boldsymbol{u}d\boldsymbol{d}^{v} , \\ 
    \end{split}
\end{equation}

Introducing the variational approximation $r({\boldsymbol{u}}|\boldsymbol{d}^{v})$ to posterior probability distribution $p({\boldsymbol{u}}|\boldsymbol{d}^{v})$, we could easily obtain the variation lower bound of ${I}\left( \boldsymbol{U}; {\boldsymbol{D}^{v}} \right)$ as follows:
\begin{equation}
\label{eq12}
    \begin{split}
         {I}\left( \boldsymbol{U}; {\boldsymbol{D}^{v}} \right)
         & \ge \iint{p({\boldsymbol{u}},\boldsymbol{d}^{v})}log\frac{r({\boldsymbol{u}}|\boldsymbol{d}^{v})}{p(\boldsymbol{u})}d\boldsymbol{u}d\boldsymbol{d}^{v}  \\ &=\iint{p({\boldsymbol{u}},\boldsymbol{d}^{v})}log{r({\boldsymbol{u}}|\boldsymbol{d}^{v})}d\boldsymbol{u}d\boldsymbol{d}^{v} + H(\boldsymbol{u}) \\ 
         & \ge \int{d\boldsymbol{d}^{v}}p({\boldsymbol{d}^{v}})\int{d{\boldsymbol{u}}p({\boldsymbol{u}}|\boldsymbol{d}^{v})log{r({\boldsymbol{u}}|\boldsymbol{d}^{v})}}. \\ 
    \end{split}
\end{equation}

Since the entropy of the $H(\boldsymbol{u})$ is non-negative and has no effect on the optimization process, it can be directly ignored. 

Similar to the calculation process of ${I}\left( \boldsymbol{U}; {\boldsymbol{D}^{v}} \right)$, for ${I}\left( \boldsymbol{U}; {\boldsymbol{C}} \right)$, we have
\begin{equation}
\label{eq13}
    \begin{split}
         {I}\left( \boldsymbol{U}; {\boldsymbol{C}} \right)
         & \ge \iint{p({\boldsymbol{u}},\boldsymbol{c})}log\frac{t({\boldsymbol{u}}|\boldsymbol{c})}{p(\boldsymbol{u})}d\boldsymbol{u}d\boldsymbol{c}  \\ 
         & \ge \int{d\boldsymbol{c}}p({\boldsymbol{c}})\int{d{\boldsymbol{u}}p({\boldsymbol{u}}|\boldsymbol{c})log{t({\boldsymbol{u}}|\boldsymbol{c})}}. \\ 
    \end{split}
\end{equation}
where $t({\boldsymbol{u}}|\boldsymbol{c})$ is the variational estimation to $p({\boldsymbol{u}}|\boldsymbol{c})$.

Bringing the above inequalities Eq.(\ref{eq10}), Eq.(\ref{eq12}), and Eq. (\ref{eq13}) into Eq.(\ref{eq7}), ${\mathcal{L}_{IB}}$ could be rewritten as
\begin{equation}
\label{eq14}
    \begin{split}
        {\mathcal{L}_{IB}}& \le\sum\limits_{v=1}^{V}\{\int{d\boldsymbol{r}^{v}}p({\boldsymbol{r}^{v}})\int{d{\boldsymbol{u}}p({\boldsymbol{u}}|\boldsymbol{r}^{v})log\frac{p({\boldsymbol{u}}|\boldsymbol{r}^{v})}{q(\boldsymbol{u})}}\\
        &-\int{d\boldsymbol{d}^{v}}p({\boldsymbol{d}^{v}})\int{d{\boldsymbol{u}}p({\boldsymbol{u}}|\boldsymbol{d}^{v})log{r({\boldsymbol{u}}|\boldsymbol{d}^{v})}}\}\\
        &- \int{d\boldsymbol{c}}p({\boldsymbol{c}})\int{d{\boldsymbol{u}}p({\boldsymbol{u}}|\boldsymbol{c})log{t({\boldsymbol{u}}|\boldsymbol{c})}}\\
    \end{split}
\end{equation}

Therefore, minimizing $\mathcal{L}_{IB}$ might be relaxed to minimize its upper bound. To further simplify and remove items that have no effect on the optimization process, we apply the Monte Carlo sampling\cite{shapiro2003monte} to approximate the integral over $\boldsymbol{r}^{v}$, $\boldsymbol{d}^{v}$, and $\boldsymbol{c}$. Then the final tractable estimation is derived as
\begin{equation}
\label{eq15}
    \begin{split}  
     {\mathcal{L}_{IB}} &\doteq 
     \frac{1}{n}\sum\limits_{i=1}^{n}\{\sum\limits_{v=1}^{V}({{\int{d{\boldsymbol{u}}p({\boldsymbol{u}}|\boldsymbol{r}_{i}^{v})log\frac{p({\boldsymbol{u}}|\boldsymbol{r}_{i}^{v})}{q(\boldsymbol{u})}}} } \\ 
    & -{\int{d{\boldsymbol{u}}p({\boldsymbol{u}}|\boldsymbol{d}_{i}^{v})log{r({\boldsymbol{u}}|\boldsymbol{d}_{i}^{v})}}}) \\  
    &-\int{d{\boldsymbol{u}}p({\boldsymbol{u}}|\boldsymbol{c}_{i})log{t({\boldsymbol{u}}|\boldsymbol{c}_{i})}}\}.\\
    \end{split}
\end{equation}

Without loss of generality, we assume the posterior probability distributions $p({\boldsymbol{u}}|\boldsymbol{d}^{v})$ and $p({\boldsymbol{u}}|\boldsymbol{c})$ obey Gaussian distribution, the mean and variance can be learned from neural networks. That is, $p({\boldsymbol{u}}|\boldsymbol{d}^{v})=\mathcal{N} (\mu_{1} (\boldsymbol{d}^{v};\phi_{1} ),\sigma_{1} (\boldsymbol{d}^{v};\phi_{1} ))$, $p({\boldsymbol{u}}|\boldsymbol{c})=\mathcal{N} (\mu_{2} (\boldsymbol{c};\phi_{2} ),\sigma_{2} (\boldsymbol{c};\phi_{2} ))$. $\phi_{1} $ and $\phi_{2}$ are parameters for the networks. Here, we utilize the reparameterization trick to get $\boldsymbol{u}=\mu_{1} (\boldsymbol{d}^{v};\phi_{1} )+\sigma_{1} (\boldsymbol{d}^{v};\phi_{1} ){\epsilon}_{1}$, $\boldsymbol{u}=\mu_{2} (\boldsymbol{c};\phi_{2}  )+\sigma_{2} (\boldsymbol{c};\phi_{2} ){\epsilon}_{2}$, where $\epsilon_{1}$ and $\epsilon_{2}$ obey standard normal distribution for benefiting backpropagation operation. As we discussed above, the objective function of minimal-sufficient unified representation learning turns out to be:
\begin{equation}
\label{eq16}
    \begin{split}
        {\mathcal{L}_{IB}}&=\frac{1}{n}\sum\limits_{i=1}^{n}\{\sum\limits_{v=1}^{V}\{{{KL\left[ p({\boldsymbol{u}}|\boldsymbol{r}_{i}^{v})||q(\boldsymbol{u}) \right]} }\\
        &-{\mathbb{E}_{\epsilon_{1}}log{r({\boldsymbol{u}}|\boldsymbol{d}_{i}^{v})}}\}-{\mathbb{E}_{\epsilon_{2}}log{t({\boldsymbol{u}}|\boldsymbol{c}_{i})}}\}.\\
    \end{split}
\end{equation}

Complementary to the above representation learning, we believe that the discriminability of unified representation, a.k.a., strong intra-cluster aggregation and inter-cluster separation, should also be considered for clustering effectiveness improvement. Therefore, we further promote $\boldsymbol{U}$ convergence to this property, unlike directly performing self-expression learning in a feature space without discriminative constraints. However, this research is clustering-oriented, and the true cluster labels are unknown. To tackle the issue, we predict pseudo-labels as additional information to aid in identifying classes. Let $\boldsymbol{l}\in \mathbb{R}^{n}$ stand for the pseudo cluster labels calculated over the consistent probability distribution $\boldsymbol{Q}$,
\begin{equation}
\label{eq17}
    \begin{split}
        {{l}_{i}}=\arg \underset{k}{\mathop{\max }}\,({\boldsymbol{q}_{i,k}})\quad k=1,2,\ldots ,K,i=1,2,\ldots ,n.\\
    \end{split}
\end{equation}

After that, we obtain a partitioned diagonal matrix $\boldsymbol{\Pi}=\{\boldsymbol{\Pi}^{k}\}_{k=1}^{K}$, in which the diagonal entry $\boldsymbol{\Pi}^{k}\in\mathbb{R}^{n\times n}$ denotes a relationship matrix corresponding to cluster $k$. ${\Pi}_{i, i}^{k}=1$ if the $i$-th sample belongs to cluster $k$, ${\Pi}_{i, i}^{k}=0$ otherwise. Since cluster assignment consistency is achieved by Eq.(\ref{eq6}) in advance, it is reasonable to estimate pseudo labels using $\boldsymbol{Q}$. Though such pseudo labels may be crude, they still function as eligible label information for obtaining $\boldsymbol{\Pi}$. In other words, $\boldsymbol{\Pi}$ can be seen that initialized relatively ideal. 

Recent work \cite{Yu2020LearningDA} presents the maximum coding rate reduction (MCR$^2$) principle. As an information-theoretic measure, it aims to learn a discriminative representation effectively by maximizing the coding rate difference between all samples and the sum of each individual cluster. Based on this principle, all features from different clusters should expand into as large a space as possible (inter-cluster separation), i.e., maximize the global coding rate. Whereas each individual cluster should be compressed into a small subspace as much as possible (intra-cluster aggregation), i.e., minimize the local coding rate. Benefiting from it, we minimize a discriminatory loss 
\begin{equation}
\label{eq18}
    \begin{split}
        {\mathcal{L}_{Dis}}
        :=-R(\boldsymbol{U})+{{R}^{c}}(\boldsymbol{U},\boldsymbol{\Pi} ),\\
    \end{split}
\end{equation}

In the above, global coding rate $R(\boldsymbol{U})$ is defined as
\begin{equation}
\label{eq19}
    \begin{split}
        R(\boldsymbol{U})\doteq \frac{1}{2} log \det (\boldsymbol{I}+\frac{d}{n{{\epsilon }^{2}}}\boldsymbol{U}{\boldsymbol{U}^{T}}),\\
    \end{split}
\end{equation}
where ${\epsilon}>0$ is a given distortion rate, ${\epsilon}^{2}$ set to $0.5$ by default, following \cite{Yu2020LearningDA}. $\det (\cdot)$ represents determinant computation function. The local coding rate is mathematically written as
\begin{equation}
\label{eq20}
    \begin{split}
        {{R}^{c}}(\boldsymbol{U},\boldsymbol{\Pi} )&\doteq \sum\limits_{k=1}^{K}{\frac{tr({{\boldsymbol{\Pi}}^{k}})}{2n}} log \det \big(\boldsymbol{I}+\big.\\
        &\big.\frac{d}{tr({\boldsymbol{\Pi}^{k}}){{\epsilon }^{2}}}\boldsymbol{U}{\boldsymbol{\Pi}^{k}}{\boldsymbol{U}^{T}}\big).\\
    \end{split}
\end{equation}

Note that when optimizing Eq.(\ref{eq18}), $\boldsymbol{\Pi}$ is fixed.

\subsubsection{Self-expression Learning Module}
Based on the unified representation $\boldsymbol{U}=\{\boldsymbol{u}_{i}\}_{i=1}^{n}$, Relation-Metric Net is designed to directly measure the similarity of all sample pairs rather than via a parameterized FC layer. See the illustration in Fig. \ref{fig2}. Formally, we generate the point-to-point self-expressive coefficient by solving the following loss function:
\begin{equation}
\label{eq21}
    \begin{split}
        {\mathcal{L}_{Rel}}&:=\sum\limits_{j=1}^{n}{\left\| {\boldsymbol{u}_{j}}-\sum\limits_{i\ne j}{s({\boldsymbol{u}_{i}},{\boldsymbol{u}_{j}},{\theta _{s}}){\boldsymbol{u}_{i}}} \right\|}_{F}^{2} \\
        &+\sum\limits_{j=1}^{n}{\sum\limits_{i\ne j}}{\left\| s({\boldsymbol{u}_{i}},{\boldsymbol{u}_{j}},{\theta _{s}}) \right\|}_{F}^{2},\\
    \end{split}
\end{equation}
where $s({\boldsymbol{u}_{i}},{\boldsymbol{u}_{j}},{\theta _{s}}):{\mathbb{R}}^{d}\times{\mathbb{R}}^{d}\to{\mathbb{R}}$ is a function implemented by the Relation-Metric Net, which has a learnable threshold parameter ${\theta }_{s}$. Meanwhile, it obviously avoids trivial solutions. The second term makes the self-expressive matrix $\boldsymbol{S}$ with the grouping effect that conforms to group highly correlated samples together\cite{Lu2012RobustAE}. In more detail, Relation-Metric Net first takes the dot product of feature representations $\boldsymbol{u}_{i}$ and $\boldsymbol{u}_{j}$, and further employs a nonlinear soft-thresholding activation ${\mathcal{T}_{{\theta }_{s}}}(\cdot)$ for effectively inducing sparsity. The self-expression learning process can be expressed as
\begin{equation}
\label{eq22}
    \begin{split}
        &s({\boldsymbol{u}_{i}},{\boldsymbol{u}_{j}},{\theta }_{s}):={\mathcal{T}_{{\theta }_{s} }}(\boldsymbol{u}_{j}^\top{\boldsymbol{u}_{i}})\\
        &\mathrm{s.t.}\text{      }{\mathcal{T}_{{\theta }_{s}}}(\boldsymbol{u}_{j}^\top{\boldsymbol{u}_{i}}):=\mathrm{sgn}(\boldsymbol{u}_{j}^\top{\boldsymbol{u}_{i}})\max (0,\left| \boldsymbol{u}_{j}^\top{\boldsymbol{u}_{i}} \right|-{{\theta }_{s}} ),        
    \end{split}
\end{equation}
where $\mathrm{sgn}(\cdot)$ represents the signum function. Through Eq.(\ref{eq22}), we can calculate a similarity score of $\boldsymbol{u}_{i}$ and $\boldsymbol{u}_{j}$, which serves as the corresponding self-expressive coefficient $\boldsymbol{s}_{ij}$ in $\boldsymbol{S}$.

Our method, in contrast to the classical approach that employs an FC layer to parameterize self-expression relations, achieves a quadratic reduction in the number of parameters and memory usage. Obviously, it is well-suited for handling arbitrary-scale datasets, including considerably large ones. Besides, we can easily observe that the self-expressive learning process is symmetric, i.e., the off-diagonal entries $\boldsymbol{s}_{ij}={\boldsymbol{s}_{ji}}$. Hence, the self-expressive coefficient matrix $\boldsymbol{S}$ enjoys the symmetric-structure property and can directly serve as the affinity matrix to apply spectral clustering.

\subsection{Parameter Optimization}
Combining all the aforementioned losses jointly into account, mathematically, the overall loss function of E$^2$MVSC has the following formula:
\begin{equation}
\label{eq2}
    \begin{split}
    {\mathcal{L}}:={\mathcal{L}_{AEs}}+{\lambda_1}{\mathcal{L}_{SS}}+{\lambda_2}{\mathcal{L}_{IB}}+{\lambda_3}{\mathcal{L}_{Dis}}
    +{\lambda_4}{\mathcal{L}_{Rel}},
    \end{split}
\end{equation}
where $\lambda_1,\lambda_2, \lambda_3, \lambda_4>0$ are trade-off parameters. 

Throughout the experiments, we first pre-train only with auto-encoders by minimizing ${\mathcal{L}_{AEs}}$, then fit the pre-trained parameters to initialize the complete model. After that, we update all parameters by minimizing objective function Eq.(\ref{eq2}) with the gradient descent algorithm. Once the network converges, the optimal self-expressive coefficient matrix $\boldsymbol{S}$ can be derived, before applying spectral clustering to get the final clustering assignments. The complete optimization procedure of E$^2$MVSC is summarized in \textbf{Algorithm} \ref{alg1}.

\begin{algorithm}
    \footnotesize
    \caption{E$^2$MVSC}
    \label{alg1}
    \LinesNumbered
    \KwIn{Multi-view data $\boldsymbol{X}^{v}$;  cluster number $K$; maximum epochs of pre-training and fine-tuning ${T}_{p}$, ${T}_{f}$; trade-off parameters ${\lambda }_{1}, {\lambda }_{2}, {\lambda }_{3}, {\lambda }_{4}$; the learning rate for pre-training and fine-tuning ${\eta}_{p}$, ${\eta}_{f}$.}
    \KwOut{Clustering result.}
        \textbf{Initialize:} view-consistent representation $\boldsymbol{C}$; unified representation $\boldsymbol{U}$; parameters ${\boldsymbol{\Theta} _{e}^{v}}$, ${\boldsymbol{\Theta} _{d}^{v}}$, ${\boldsymbol{\Theta}_{clu}}$, and ${\theta }_{s}$ with random values.\\
        \emph{\# Step1: Pre-training for Auto-Encoders.}\\
        \While{not converged or not reach ${T}_{p}$}{
        Pre-train auto-encoder of each view by minimizing Eq.(\ref{eq4})\;
        Compute the gradient of Eq.(\ref{eq4}) and update ${\boldsymbol{\Theta} _{e}^{v}}$, ${\boldsymbol{\Theta} _{d}^{v}}$, and $\boldsymbol{C}$. \\
        }
        Obtain pre-trained parameters ${\boldsymbol{\Theta} _{e}^{v}}$, ${\boldsymbol{\Theta} _{d}^{v}}$, and $\boldsymbol{C}$.\\
        \emph{\# Step2: Fine-tuning for entire model.}\\
        \While{not converged or not reach ${T}_{f}$}{
        Train the complete network using Eq.(\ref{eq2}) on data $\boldsymbol{X}^{v}$\;
        Update network parameters ${\boldsymbol{\Theta} _{e}^{v}}$, ${\boldsymbol{\Theta} _{d}^{v}}$, ${\boldsymbol{\Theta}_{clu}}$, ${\theta }_{s}$ and $\boldsymbol{S}$ through gradient descent to minimize the objective in Eq.(\ref{eq2}). \\
        }
        Obtain self-expressive coefficient matrix $\boldsymbol{S}$. \\
    Perform spectral clustering algorithm on matrix $\boldsymbol{S}$.
    \end{algorithm}

\subsection{Connections to Related Works}

E$^2$MVSC has two main purposes: deriving a unified representation for clustering that is minimal-sufficient and discriminative, and decoupling the parameter scale from sample numbers to quickly yield self-expressive coefficients. Although learning a minimal-sufficient representation seems to be a relatively common topic in previous work\cite{Wan2021MultiViewIR, Wang2022SelfSupervisedIB, Yan2023MultiviewSC}, it should be pointed out that, our E$^2$MVSC differs significantly from such methods. To be concrete, existing methods based on information bottleneck theory, such as MIB \cite{DBLP:conf/iclr/Federici0FKA20}, and MSCIB \cite{Yan2023MultiviewSC}, focus on learning a representation that captures consistent information, without considering the complementarity among multi-view data. This is mainly because, in these methods, the information not shared across views is enforced as clustering-irrelevant and discarded, which undoubtedly loses valuable complementary information. 

MFLVC\cite{XuT0P0022} proposes a multi-level learning framework, where view-private information is implicitly filtered out during high-level feature learning. In contrast, our E$^2$MVSC explicitly decouples consistent, complementary, and superfluous information from multiple views by introducing a novel constraint of soft clustering assignment similarity. Then our method, maximizing the mutual information between unified representation and complementary representations, and minimizing the mutual information with superfluous representations, can sufficiently guarantee the preservation of complementarity in unified representation. Besides, we consider and enhance the discriminability of unified representation. 

SENet\cite{ZhangYVL21} finds a simplistic metric-based network that produces self-expressive coefficients, without the network parameters needed to scale with sample numbers. Though this is the same starting point as our work, we extend metric learning to the multi-view learning setting and, moreover, directly generate a symmetric self-expressive matrix for spectral clustering.

\section{Experiments}
\label{Experiments}
In this section, we conduct comprehensive experiments on multiple publicly available multi-view datasets, alongside comparing E$^2$MVSC with several state-of-the-art related works, to demonstrate the goodness of the proposed method.

\subsection{Experimental Settings}

\subsubsection{Multi-view Datasets}
We select seven different image types of benchmark datasets and perform extensive experiments on them, including \textbf{Hand Written}\footnote{\url{https://archive.ics.uci.edu/dataset/72/multiple+features}}, \textbf{Caltech101-20}\footnote{\url{http://www.vision.caltech.edu/Image_Datasets/Caltech101}}, \textbf{CCV}\footnote{\url{https://www.ee.columbia.edu/ln/dvmm/CCV/}}, \textbf{Caltech101-all}, \textbf{SUNRGBD}\footnote{\url{http://rgbd.cs.princeton.edu/}}, \textbf{NUS-WIDE-Object}\footnote{\url{https://lms.comp.nus.edu.sg/wp-content/uploads/2019/research/nuswide/NUS-WIDE.html}}, and \textbf{YouTubeFace}\footnote{\url{https://www.cs.tau.ac.il/~wolf/ytfaces/}}. Among them, sample numbers vary from 2,000 to 101,499, with different numbers of views, to sufficiently evaluate the scalability of our E$^2$MVSC. All datasets follow the feature extraction settings in previous works\cite{Wang2021FastPM, Liu2021OnepassMC}. Concretely, the specifications of these datasets are shown in Table \ref{tab2}. 

\begin{table*}[!t]
\centering
\caption{Description of Datasets Used in Experiments}
\label{tab2}
\footnotesize
\renewcommand\arraystretch{1.3}
\tabcolsep 5.5pt 
\begin{tabular*}{\textwidth}{l|ccccc}
\toprule
 Datasets  & $\textrm{\#}$ Total samples & $\textrm{\#}$ Clusters & $\textrm{\#}$ Views & $\textrm{\#}$ Feature Dimensions & Category  \\
 \midrule
    Hand Written      & 2,000 & 10 & 6 &  FOU(76),FAC(216),KAR(64),PIX(240),ZER(47),MOR(6)          & Digital Image   \\
    Caltech101-20     & 2,386 & 20 & 6 &  Gabor(48),WM(40), CENT(254),HOG(1,984),GIST(512),LBP(928) & Object Image     \\
    CCV               & 6,773 &20  & 3 & STIP(20),SIFT(20),MFCC(20)                                 & Video Event      \\
    Caltech101-all    & 9,144 & 102 & 5 &  Gabor(48),WM(40), CENT(254),GIST(512),LBP(928)           & Object Image      \\
    SUNRGBD           & 10,335 & 45 & 2 & View1(4,096), View2(4,096)                                & Scene Image         \\
    NUS-WIDE-Object   & 30,000 & 31  & 5 & CH(65), CM(226), CORR(145), EDH(74), WT(129)             &  Object Image            \\
    YouTubeFace       & 101,499 & 31  & 5 &  View1(64),View2(512),View3(64),View4(647),View5(838)   & Face Video                \\
     
\bottomrule
\end{tabular*}
\end{table*}

\subsubsection{Compared Methods}
To better verify the superiority and competitiveness of our method, we compare E$^2$MVSC with ten representative related works from three perspectives. Including a large-scale single-view SC method, \textbf{SENet} \cite{ZhangYVL21}, and five prominent deep multi-view clustering methods, \textbf{RMSL}\cite{Li2019ReciprocalMS}, \textbf{MIB} \cite{DBLP:conf/iclr/Federici0FKA20}, \textbf{DMSC-UDL}\cite{Wang2021DeepMS}, \textbf{MFLVC}\cite{XuT0P0022}, and \textbf{SDMVC}\cite{Xu2023SelfSupervisedDF}. In addition, four SOTA large-scale MVSC methods are also used here for comparison, namely \textbf{LMVSC}\cite{KangZZSHX20}, \textbf{SMVSC}\cite{Sun2021ScalableMS}, \textbf{FPMVS-CAG}\cite{Wang2021FastPM}, and \textbf{EOMSC-CA}\cite{Liu2022EfficientOM}. 

\subsubsection{Implementation Details}
In our model, the encoder structure is composed of two-layer fully connected neural networks (${d}_{v}$-200-$d$), and two output heads have the same dimension $d=20$. Each fully connected layer is followed by a sigmoid layer to perform non-linear activation. The decoder is symmetric to the corresponding encoder, yet the input dimension changes to 60. We use Adam optimizer for optimization with $10^{-3}$ and $10^{-4}$ learning rates in pre-training and fine-tuning processes, respectively. The maximum epochs for pre-training and fine-tuning are set at 1,000 and 300, respectively. The batch size is set to the input data size. Additionally, we adopt the grid search strategy to find optimal trade-off parameters. In the experiment, $\lambda_1,\lambda_2,\lambda_3,\lambda_4$ are all set to 1. Elaborate on parameter sensitivity analysis of our E$^2$MVSC can refer to Section \ref{subsectionD}. 

For a fair comparison, we download demo codes of all competitors from the respective authors’ websites and fine-tune the hyper-parameters based on suggestions specified in the original papers to present optimal performance. Note that SENet was performed on each individual view, respectively, and the best clustering result is reported, termed SENet-Best. For DMSC-UDL, we replace its convolutional autocoders with fully connected networks consistent with our model to process feature views. 

E$^2$MVSC and deep competitors are implemented by PyTorch on a desktop computer equipped with an Intel(R) Core(TM) i9-10900F CPU, one NVIDIA Geforce RTX 3080ti GPU, and 32GB RAM. Other traditional comparison methods are performed under the same environment with Matlab R2020a. The source code and datasets of E$^2$MVSC will be released on GitHub upon acceptance.

\subsubsection{Evaluation Metrics}
There are four commonly used evaluation metrics employed to comprehensively measure effectiveness, including Accuracy (ACC), Normalized Mutual Information (NMI), Purity (PUR), and Fscore, each of which favors different properties of clustering. The detailed definitions of these metrics can be found in \cite{Li2020MultiviewCA}. The higher the value, the better the performance. To make experimental results more reliable, all methods are run ten times independently, giving the mean values of every metric over all datasets. 
 
\subsection{Experimental Results}
\subsubsection{Clustering Results Analysis}
We conducted experiments with the proposed method versus ten comparison methods on seven datasets. Table \ref{tab4} summarizes clustering results regarding four evaluation metrics, and several fundamental observations could be made from it. 

\begin{itemize}
    \item Our E$^2$MVSC achieves the best performance among all competitors in most cases, demonstrating the effectiveness of the proposed method. As sample numbers increase, clustering becomes more difficult. This may be attributed to the increase in sample sizes introducing more noise information that interferes with clustering, increasing the probability of misclassification occurring. Even so, E$^2$MVSC is always in the leading position. For example, on SUNRGBD, our method largely outperforms the second-best method, FPMVS, raising the performance by around 4.84\%, 6.84\%, and 9.89\% in terms of ACC, NMI, and Fscore, respectively.  

    \item As observed on small-scale datasets (less than 9,000 samples), deep-based methods generally produce higher performance than anchor-based ones. Such as, compared with EOMSC, 15.47\% ACC and 28.01\% NMI improvements can be achieved by our E$^2$MVSC on Caltech101-20. This, on the one hand, demonstrates that deep learning is helpful to better fit complex data distributions and extract a higher-quality representation than traditional machine learning methods. On the other hand, it verifies that anchor-based methods inevitably lose some accuracy in reducing time consumption, as discussed earlier.

    \item For large-scale datasets, deep-based methods are instead overall inferior to anchor-based ones. One possible reason is that anchor-based methods, selecting anchor points to approximate the overall data distribution, are robust to increasing noise information. Moreover, deep methods are not designed specifically for large-scale data and lack certain constraints. Especially RMSL and DMSC-UDL, which use the FC layer to conduct self-expression, easily suffer from the out-of-memory issue. In contrast, E$^2$MVSC is feasible on large-scale data benefiting from the devised Relation-Metric Net. Besides, our method reduces interference from redundant information by explicitly decoupling multi-view information.

    \item It is worth noting that, in some datasets (e.g., Caltech101-all, SUNRGBD), the single-view clustering method, SENet, outperforms most multi-view ones. This phenomenon suggests that complementary and consistent information among views, once not well explored and exploited, can degrade performance. Although DMSC-UDL enhances diversity across views, it is difficult to preserve consistency. MIB and MFLVC enforce the consistency of multiple views, while important complementary information is not well utilized, leading to clustering effects that are less effective. Differently, our method could learn a high-quality unified representation that fully integrates consistent and complementary information.

\end{itemize}

The above analysis demonstrates the favorable scalability and stability of our method, which is not only applicable to clustering large-scale datasets but also performs well on various types of datasets.
 
\begin{table*}[!t]
\centering
\caption{Performance (ACC, NMI, PUR, Fscore) comparison of different clustering methods on seven datasets (mean \%). The best and second-best results are highlighted in bold \color{red}{\textbf{red}} / \color{blue}{\textbf{blue}}. \color{black}{The symbol “N/A” means out of memory or out of time.}}
\label{tab4}
\footnotesize
\renewcommand\arraystretch{1.3}
\tabcolsep 4.7pt 
\begin{tabular*}{\textwidth}{ccccccccccccc}  
\toprule

    \multirow{2}*{Datasets} &\multirow{2}*{Metrics}  &SENet-Best  & {RMSL}  & DMSC-UDL & MIB &MFLVC &SDMVC & LMVSC & SMVSC &FPMVS  &EOMSC & E$^2$MVSC\\ \cmidrule(r){3-3}   \cmidrule(r){4-5}  \cmidrule(r){6-8}  \cmidrule(r){9-12} \cmidrule(r){13-13} 
    ~  & ~ &(2021)& (2019) & (2021) & (2020) & (2022) & (2023) &(2020) &(2021) &(2022) &(2022) & (Ours)\\\midrule
    \cellcolor{gray!20}~    &ACC    & 87.75    & 83.75    &93.90   & 83.88    & 80.75    & \color{blue}{\textbf{96.95}}    & 91.30  & 82.05     & 82.85     & 76.00     & \color{red}{\textbf{97.30}}    \\
                                  \cellcolor{gray!20}~ &NMI    & 80.65    & 73.46    &88.03   & 74.00    & 84.38    & \color{blue}{\textbf{93.13}}    & 83.24  & 78.86     & 77.29     & 82.08     & \color{red}{\textbf{93.68}}    \\
                                  \cellcolor{gray!20}~ &PUR    & 87.75    & 83.75    &93.90   & 83.88    & 83.43    & \color{blue}{\textbf{96.95}}    & 91.30  & 82.05     & 82.85     & 76.20     & \color{red}{\textbf{97.30}}    \\
    \cellcolor{gray!20}\multirow{-4}*{Hand Written}     &Fscore & 79.60    & 72.30    &88.62   & 72.73    & 81.52    & \color{blue}{\textbf{94.08}}    & 84.42  & 75.24     & 76.04     & 73.42     & \color{red}{\textbf{94.75}}    \\  \midrule
    \cellcolor{gray!20}~    &ACC    & 50.33    & 35.25    &49.14   & 33.09    & 44.51    & \color{blue}{\textbf{71.58}}    & 52.03  & 64.04     & 64.83     & 66.05     & \color{red}{\textbf{81.52}}    \\
                                  \cellcolor{gray!20}~ &NMI    & 60.29    & 47.82    &66.31   & 42.40    & 56.63    & \color{blue}{\textbf{71.76}}    & 51.19  & 58.65     & 61.37     & 50.46     & \color{red}{\textbf{78.47}}    \\
                                  \cellcolor{gray!20}~ &PUR    & 76.94    & 65.97    &\color{blue}{\textbf{80.84}}  & 64.30    & 72.51  & 78.79     & 55.58     & 69.90    & 71.41     & 66.39    & \color{red}{\textbf{83.11}}    \\
                                  \cellcolor{gray!20}\multirow{-4}*{Caltech101-20} &Fscore & 54.05    & 38.30    &55.50   & 33.78    & 63.16    &  \color{blue}{\textbf{72.24}}    & 52.10  & 64.97     & 68.39     & 61.50    & \color{red}{\textbf{79.26}}    \\  \midrule
                                  
    \cellcolor{gray!20}~              &ACC    & 19.87    & 18.53    &21.11   & 13.95    & \color{blue}{\textbf{22.38}}    & 20.70    & 14.52  & 21.14     & 20.91     & 22.01     & \color{red}{\textbf{22.84}}    \\
                                  \cellcolor{gray!20}~ &NMI    & 17.44    & 14.99    &16.48   & 09.19    & \color{red}{\textbf{19.43}}    & \color{blue}{\textbf{19.36}}    & 10.01  & 15.85     & 16.28     & 17.55     & 17.86    \\
                                  \cellcolor{gray!20}~ &PUR    & 22.97    & 22.46    &23.77   & 17.14    & \color{blue}{\textbf{25.07}}   & 24.31    & 17.93  & 23.46     & 24.95     & 24.99     & \color{red}{\textbf{25.10}}    \\
                                  \cellcolor{gray!20}\multirow{-4}*{CCV} &Fscore & 12.17    & 11.37    &12.71   & 09.77    & \color{blue}{\textbf{14.13}}     & 13.08    & 09.37  & 13.83     & 12.63     & 13.27     & \color{red}{\textbf{14.68}}    \\ \midrule
                                  
    \cellcolor{gray!20}~   &ACC    & 23.49    & N/A      &22.62   & 11.58    & 15.57    & 21.03    & 11.65  & 29.04     & \color{blue}{\textbf{29.81}}     & 22.32     & \color{red}{\textbf{38.37}}    \\
                                  \cellcolor{gray!20}~ &NMI    & 42.39    & N/A      &\color{blue}{\textbf{46.10}}   & 28.02    & 37.56    & 41.17    & 25.30  & 35.03     & 38.14     & 24.70     & \color{red}{\textbf{48.84}}    \\
                                  \cellcolor{gray!20}~ &PUR    & 39.26    & N/A      &\color{blue}{\textbf{45.42}}   & 26.32    & 23.90    & 36.63    & 22.83  & 33.32     & 36.12     & 25.12     & \color{red}{\textbf{46.02}}    \\
                                  \cellcolor{gray!20}\multirow{-4}*{Caltech101-all} &Fscore & 20.62    & N/A      &22.56   & 08.38    & 13.28    & 19.08    & 05.44  & 20.07     &  \color{blue}{\textbf{22.89}}     & 10.83     & \color{red}{\textbf{38.12}}    \\  \midrule
                                  
    \cellcolor{gray!20}~          &ACC    & 18.79    & N/A      &15.71   & 10.34    & 11.16    & 19.11    & 18.25  & 21.85     & \color{blue}{\textbf{24.36}}     & 23.95     & \color{red}{\textbf{29.20}}    \\
                                  \cellcolor{gray!20}~ &NMI    & \color{blue}{\textbf{26.83}}    & N/A      &21.12   & 04.81    & 14.27    & 22.81    & 25.46  & 22.72     & 22.21     & 23.75     & \color{red}{\textbf{29.05}}    \\
                                  \cellcolor{gray!20}~ &PUR    & \color{blue}{\textbf{40.29}}    & N/A      &33.70   & 14.35    & 24.01    & 33.33    & 37.71  & 32.56     & 31.07     & 35.20     & \color{red}{\textbf{40.96}}    \\
                                  \cellcolor{gray!20}\multirow{-4}*{SUNRGBD} &Fscore & 13.69    & N/A      &10.99   & 09.45    &  08.93   & 13.77    & 11.56  & 14.68     & \color{blue}{\textbf{15.68}}     & 14.94     & \color{red}{\textbf{23.07}}    \\ \midrule
                                  
    \cellcolor{gray!20}~         &ACC    & 12.01    & N/A      & N/A    & 10.71    & 16.29    & 17.49    & 14.35  & \color{blue}{\textbf{19.11}}     & 19.12     & 18.89     & \color{red}{\textbf{20.79}}    \\
                                  \cellcolor{gray!20}~ &NMI    & 11.11    & N/A      & N/A    & 09.88    & \color{blue}{\textbf{18.02}}    & 11.75    & 12.22  & 12.75     & 13.16     & 13.44     & \color{red}{\textbf{18.42}}    \\
                                  \cellcolor{gray!20}~ &PUR    & 23.95    & N/A      & N/A    & 21.31    & \color{red}{\textbf{28.16}}   & 22.96    & 19.14  & 23.50     & 23.33     & 24.06     & \color{blue}{\textbf{27.35}}    \\
                                  \cellcolor{gray!20}\multirow{-4}*{NUS-WIDE} &Fscore & 08.02    & N/A      & N/A    & 07.74    & 12.31     & 13.41    & 09.31  & 12.57     & 13.12     & \color{blue}{\textbf{13.69}}     & \color{red}{\textbf{22.10}}    \\ \midrule
                                  
    \cellcolor{gray!20}~      &ACC    & 17.50    & N/A      & N/A    & 14.89    & 23.32    & \color{blue}{\textbf{25.82}}    & 12.95  & 25.41     & 23.38     & 24.80     & \color{red}{\textbf{29.27}}    \\
                                  \cellcolor{gray!20}~ &NMI    & 16.16    & N/A      & N/A    & 14.11    & \color{red}{\textbf{26.58}}    & 23.95    & 05.28  & 24.30     & 23.48     & 02.02     & \color{blue}{\textbf{24.61}}    \\
                                  \cellcolor{gray!20}~ &PUR    & 28.62    & N/A      & N/A    & 28.43    & 33.52    & \color{blue}{\textbf{34.75}}    & 26.80  & 32.78     & 31.30     & 26.64     & \color{red}{\textbf{38.18}}    \\
                                  \cellcolor{gray!20}\multirow{-4}*{YouTubeFace} &Fscore & 11.88    & N/A      & N/A    & 11.61    & \color{blue}{\textbf{21.15}}    & 18.00    & 11.01  & 13.67     & 13.61     & 16.69     & \color{red}{\textbf{28.91}}    \\ 
   
\bottomrule
\end{tabular*}
\end{table*}

\subsubsection{Visualization Analysis}


\begin{figure*}[!t]
\centering
\includegraphics[width=\linewidth]{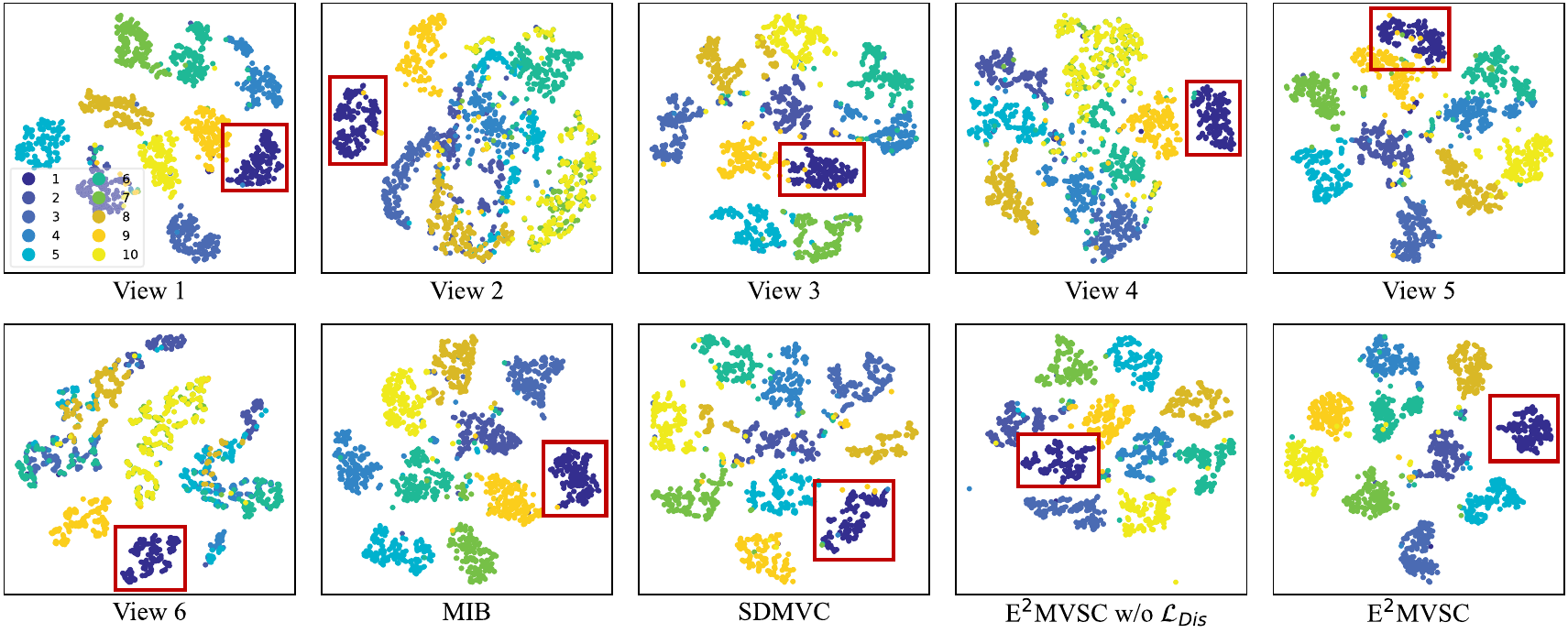}
\caption{The t-SNE visualization comparison of learned feature representations using various methods. Take the Hand Written dataset for example.} 
\label{fig4}
\end{figure*}

To further illustrate that learned unified representation possesses discriminability, we employ t-distributed Stochastic Neighbor Embedding (t-SNE) to visualize raw views and unified feature representation learned by different methods (MIB, SDMVC, and E$^2$MVSC). Notably, E$^2$MVSC, which excludes the discriminative constraint, is denoted as E$^2$MVSC w/o $\mathcal{L}_{Dis}$. Taking the Hand Written dataset as an example here, visualization results are shown in Fig. \ref{fig4}, in which different colors indicate different clusters.

The results reveal that the clustering structure gained with unified representation shows a pronounced visible boundary among classes, but raw features without processing are cluttered. Especially, E$^2$MVSC recovers cluster structure with better inter-cluster separability and intra-cluster compactness, surpassing other methods without considering discriminability (the red rectangle highlights an example cluster). Besides, using our method, only a few samples located at edges are divided incorrectly, unlike others where more sample points from different clusters are mixed together. These findings emphasize the significance of incorporating the MCR$^2$ principle to pursue discrimination and further support the goodness of our method.

\subsubsection{Efficiency and Computational Cost Analysis} For efficiency evaluation, we devise a variant of our model that replaces the Relation-Metric Net with an FC layer to compute the self-expressive coefficients, like most existing methods, namely E$^2$MVSC w FC. We calculate the training time, model parameters (params), floating point operations (FLOPs), and clustering accuracy of E$^2$MVSC and its variant on four different scale datasets. Note that as both two models share the same pre-training processing, only the fine-tuning time is reported here for comparison.

From the results in Table \ref{tab5}, it can be seen that the proposed method extremely reduces computational cost. For example, on the YouTubeFace dataset, E$^2$MVSC has 120.78 GFLOPs and only 1.19M parameters, while E$^2$MVSC w FC reaches 738.91 GFLOPs and 10.30B parameters. This is because we decouple the parameter scale of the self-expression learning network from sample numbers. Besides, our method also reduces the training time by around 50\% and brings varying degrees of gain to clustering performance. In brief, E$^2$MVSC is effective and efficient for clustering tasks and especially exhibits a significant advantage for large-scale datasets.

\begin{table}[!h]
\centering
\caption{Efficiency and computational cost analysis related to Time (s), Parameters, FLOPs, and ACC (\%). The best results are marked in \textbf{bold}. Cal-20 is Caltech101-20 for short.}
\label{tab5}
\footnotesize
\renewcommand\arraystretch{1.4}
\tabcolsep 4pt 
\resizebox{\linewidth}{!}{
\begin{tabular}{cccccc}
\toprule
    \multirow{2}*{Metrics}  &  \multirow{2}*{Methods}  & \multicolumn{4}{c}{Datasets}           \\\cmidrule{3-6}
     ~ & ~  & Hand Written   & Cal-20  & NUS-WIDE   & YouTubeFace                                    \\\midrule
    Time & E$^2$MVSC w FC   & 226.11     & 399.05     &  N/A     &  N/A          \\
    {$\downarrow$}&  \cellcolor{gray!20}E$^2$MVSC       &\cellcolor{gray!20}\textbf{142.42}     &\cellcolor{gray!20}\textbf{188.03}     &\cellcolor{gray!20}\textbf{1272.22}    &\cellcolor{gray!20}\textbf{3331.83}             \\\midrule
    Params & E$^2$MVSC w FC   & 4.67M     & 7.62M     & 0.90B     & 10.30B     \\
    $\downarrow$ &\cellcolor{gray!20}E$^2$MVSC          &\cellcolor{gray!20}\textbf{0.67M}     &\cellcolor{gray!20}\textbf{1.92M}     &\cellcolor{gray!20}\textbf{0.60M}     &\cellcolor{gray!20}\textbf{1.19M}                \\\midrule
    FLOPs & E$^2$MVSC w FC   & 1.58G     & 4.91G     & 71.87G     & 738.91G   \\
    $\downarrow$ &\cellcolor{gray!20}E$^2$MVSC          &\cellcolor{gray!20}\textbf{1.34G}     &\cellcolor{gray!20}\textbf{4.57G}     &\cellcolor{gray!20}\textbf{17.87G}     &\cellcolor{gray!20}\textbf{120.78G}                \\\midrule
    ACC & E$^2$MVSC w FC  & 97.25     & 59.43     & N/A     & N/A            \\
    $\uparrow$ & \cellcolor{gray!20} E$^2$MVSC      & \cellcolor{gray!20}\textbf{97.30}     & \cellcolor{gray!20}\textbf{81.52}  & \cellcolor{gray!20}\textbf{20.79}     & \cellcolor{gray!20}\textbf{29.27}   \\
\bottomrule
\end{tabular} }
\end{table}

\subsection{Model Analysis}


\subsubsection{Whether Information Bottleneck is Necessary}
E$^2$MVSC learns a unified representation following information bottleneck theory, aiming to not only preserve clustering-relevant information but also eliminate irrelevant information nuisances. To verify its contribution to boosting clustering effects, this subsection empirically studies the other three widely used feature fusion algorithms on our proposed model, including concatenation, fusion, and adaptive fusion. For concatenation, we directly concatenate consistent and complementary representations as the unified representation. Fusion operation involves arithmetic averaging. Moreover, we develop a learnable adaptive fusion layer to integrate consistent and complementary information. Results on the Hand Written, CCV, and SUNRGBD datasets are reported in Table \ref{tab6}. It can be found that our model based on IB theory attains more promising clustering performance, while other methods perform inconsistently on different datasets. 

\begin{table}[!t]
\centering
\caption{Comparison with Different Unified Representation Learning Methods. The best results are marked in bold \color{red}{\textbf{red}}.}
\label{tab6}
\footnotesize
\renewcommand\arraystretch{1.3}
\tabcolsep 8 pt 
\begin{tabular}{lccccc} 
\toprule
    Metrics &   Datasets & Con.  &Fus. & A-Fus. & Ours    \\ 
\midrule
    \multirow{3}*{ACC}      & Hand Written  & 95.95 & 80.65  & 96.00   & \color{red}{\textbf{97.30}} \\
    ~                       & CCV           & 21.18 & 20.46  & 20.31   & \color{red}{\textbf{22.84}} \\
    ~                       & SUNRGBD       & 23.60 & 23.67  & 25.19   & \color{red}{\textbf{29.20}} \\
    \hline
    \multirow{3}*{NMI}     & Hand Written  & 92.25 & 81.92         & 92.47   & \color{red}{\textbf{93.68}} \\
     ~                      & CCV           & 17.71 & 16.15 & 17.20   & \color{red}{\textbf{17.86}} \\
    ~                       & SUNRGBD       & 26.50 & 26.35          & 27.26   & \color{red}{\textbf{29.05}} \\
    \hline
    \multirow{3}*{PUR}     & Hand Written  & 95.89 & 84.05  & 95.92    & \color{red}{\textbf{97.30}}\\
     ~                      & CCV           & 25.10 & 24.98   & 25.06  & \color{red}{\textbf{25.10}} \\
    ~                       & SUNRGBD       & 39.15 & 38.28   & 39.85  & \color{red}{\textbf{40.96}} \\
    \hline
    \multirow{3}*{Fscore}  & Hand Written  & 94.13 & 79.43  & 94.24    & \color{red}{\textbf{94.75}}\\
     ~                      & CCV           & 13.37 & 13.99   & 13.54  & \color{red}{\textbf{14.68}} \\
    ~                       & SUNRGBD       & 17.05 & 18.41   & 18.49  & \color{red}{\textbf{23.07}} \\

\bottomrule
\end{tabular}
\end{table}
 
\subsubsection{Ablation Study}

\begin{table*}[!t]
\centering
\caption{Clustering performance (NMI\%) of E$^2$MVSC with different losses on all datasets. The best results are marked in \textbf{Bold}.}
\label{tab7}
\footnotesize
\renewcommand\arraystretch{1.5}
\tabcolsep 0.015\linewidth 
\resizebox{0.8\textwidth}{!}{
\begin{tabular}{lccccccc}
\toprule
    Method &   Hand Written & Caltech101-20 & CCV &  Caltech101-all  & SUNRGBD & NUS-WIDE-Object  & YouTubeFace\\\hline
    E$^2$MVSC w/o $\mathcal{L}_{ortho}$    &   91.24 & 76.30 & 17.22 &  45.92  & 26.17 & 17.67  & 23.23\\
    E$^2$MVSC w/o $\mathcal{L}_{SS}$       &   90.37 & 74.72 & 17.77 &  45.53  & 25.90 & 14.61  & 21.17\\
    E$^2$MVSC w/o $\mathcal{L}_{Dis}$      &   92.45 & 76.00 & 17.81 &  46.48  & 26.35 & 16.57  & 21.94\\
   \rowcolor{gray!20} E$^2$MVSC                              &   \textbf{93.68} & \textbf{78.47} & \textbf{17.86} &  \textbf{48.84}  & \textbf{29.05} &\textbf{18.42}  &  \textbf{24.61}\\
\bottomrule
\end{tabular}  }
\end{table*}

In addition, to better understand the rationality and effectiveness of each loss component in E$^2$MVSC, we carried out a set of detailed ablation studies on all seven datasets. Specifically, since both the within-view reconstruction loss and the self-expression learning loss are essential for our model, we formulate three variants in which one of the remaining loss components is removed in turn. E$^2$MVSC with all loss components is set as the baseline, and three variants are as follows: E$^2$MVSC w/o $\mathcal{L}_{ortho}$, indicates the objective functions Eq.(\ref{eq4}) and Eq.(\ref{eq2}) without the orthogonal constraint. E$^2$MVSC w/o $\mathcal{L}_{SS}$ is optimized without the cluster assignment similarity constraint, whose $\lambda_2$ is set to zero. E$^2$MVSC w/o $\mathcal{L}_{Dis}$, refers to the one excluding discriminative constraint, and set $\lambda_3=0$.

The meaning of these loss components was thoroughly discussed in Section \ref{subsectionC}, and further experimental verification is provided in Table \ref{tab7}. It is evident from the results that, in comparison with three variants, E$^2$MVSC, coupling all loss terms, consistently achieves the best results (in metrics of NMI) across seven datasets. When one of the crucial modules is removed, the clustering effect degrades. It strongly turns out that each loss component is reasonable and effectively contributes to clustering performance.

\begin{figure*}[!t]
\centering
	\subfigcapskip=-3pt 
\subfigure[HW]{
		\includegraphics[height=0.082\textheight,width=2.27cm]{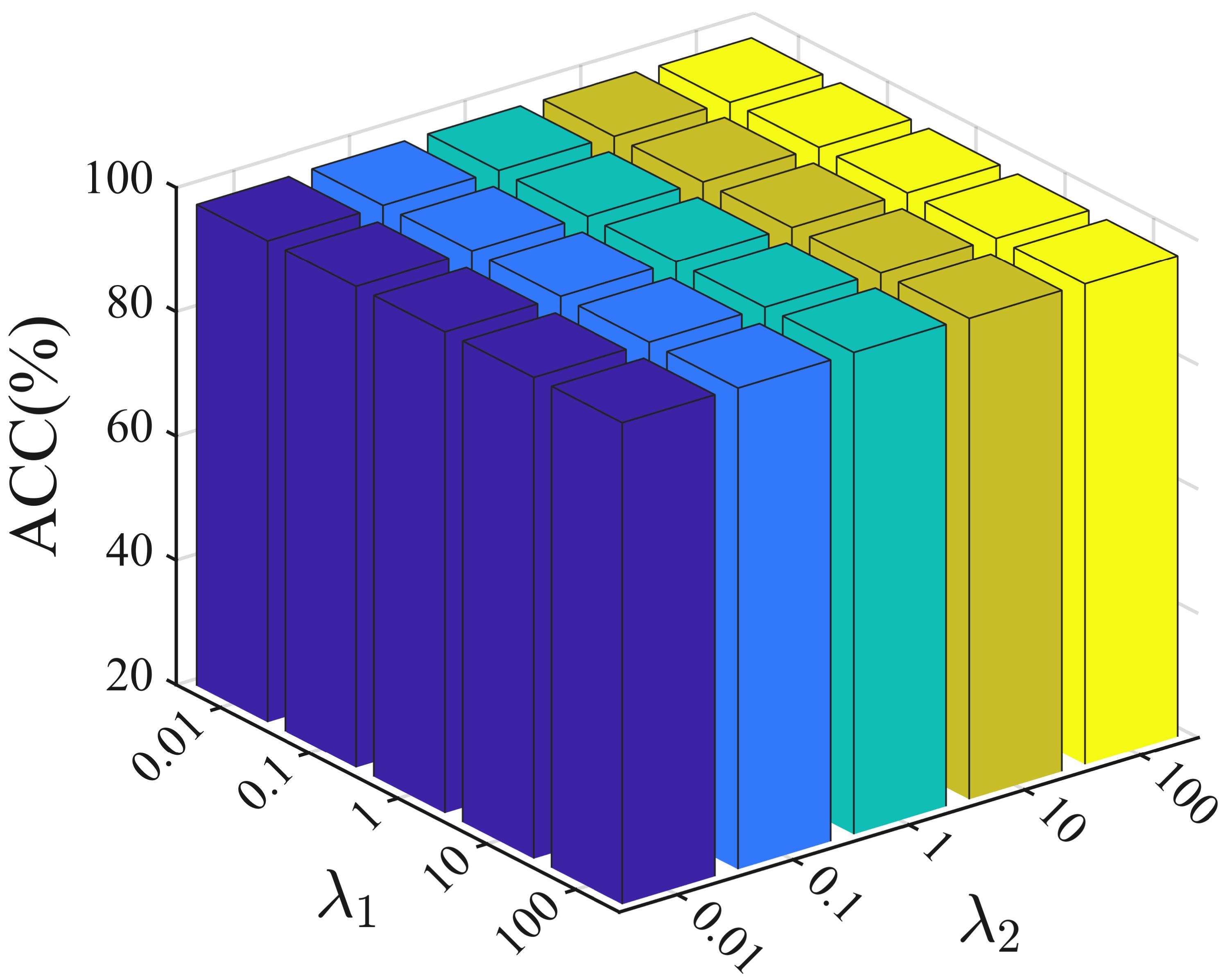}}
\subfigure[Caltech101-20]{
		\includegraphics[height=0.082\textheight,width=2.27cm]{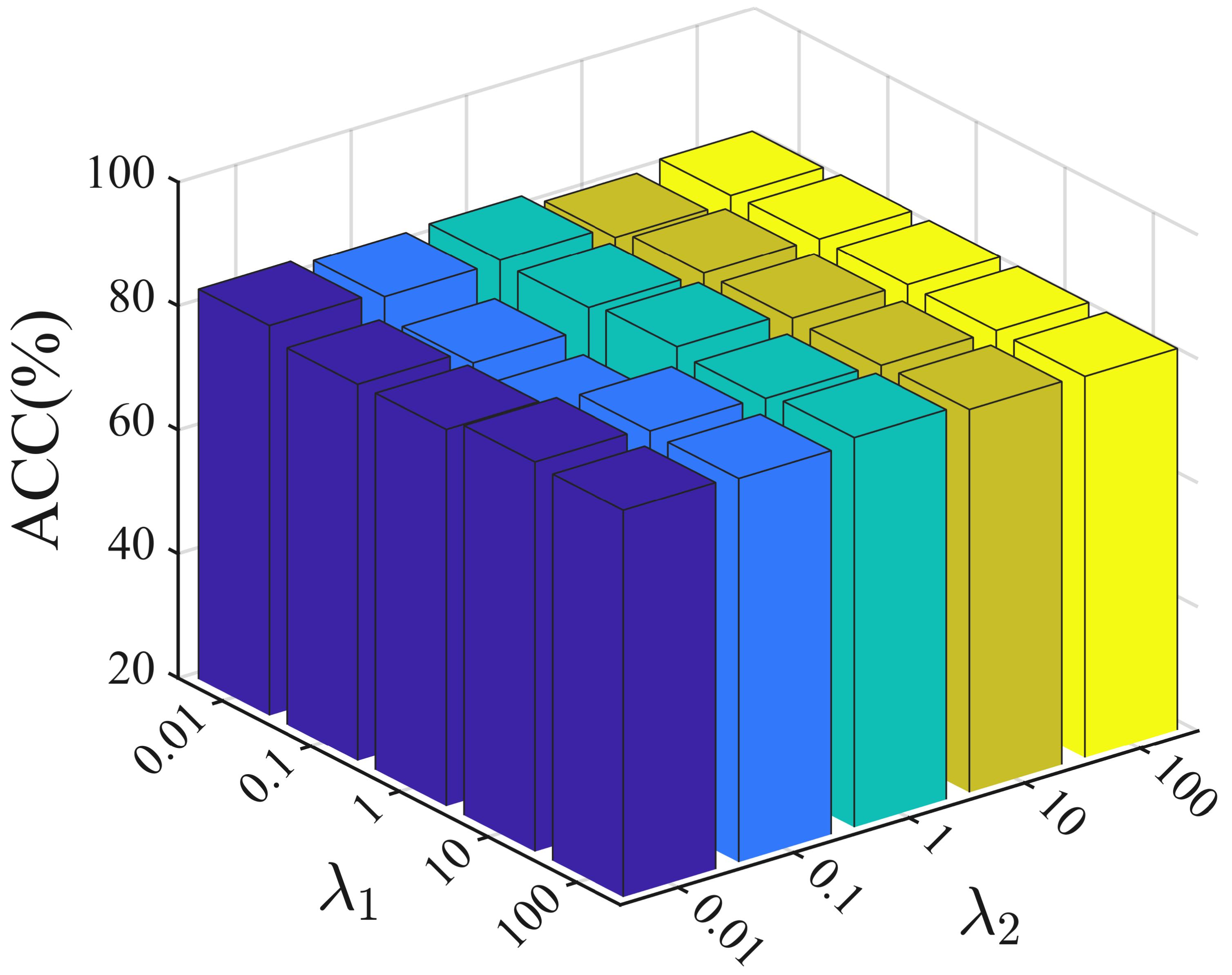}}
\subfigure[CCV]{
		\includegraphics[height=0.082\textheight,width=2.27cm]{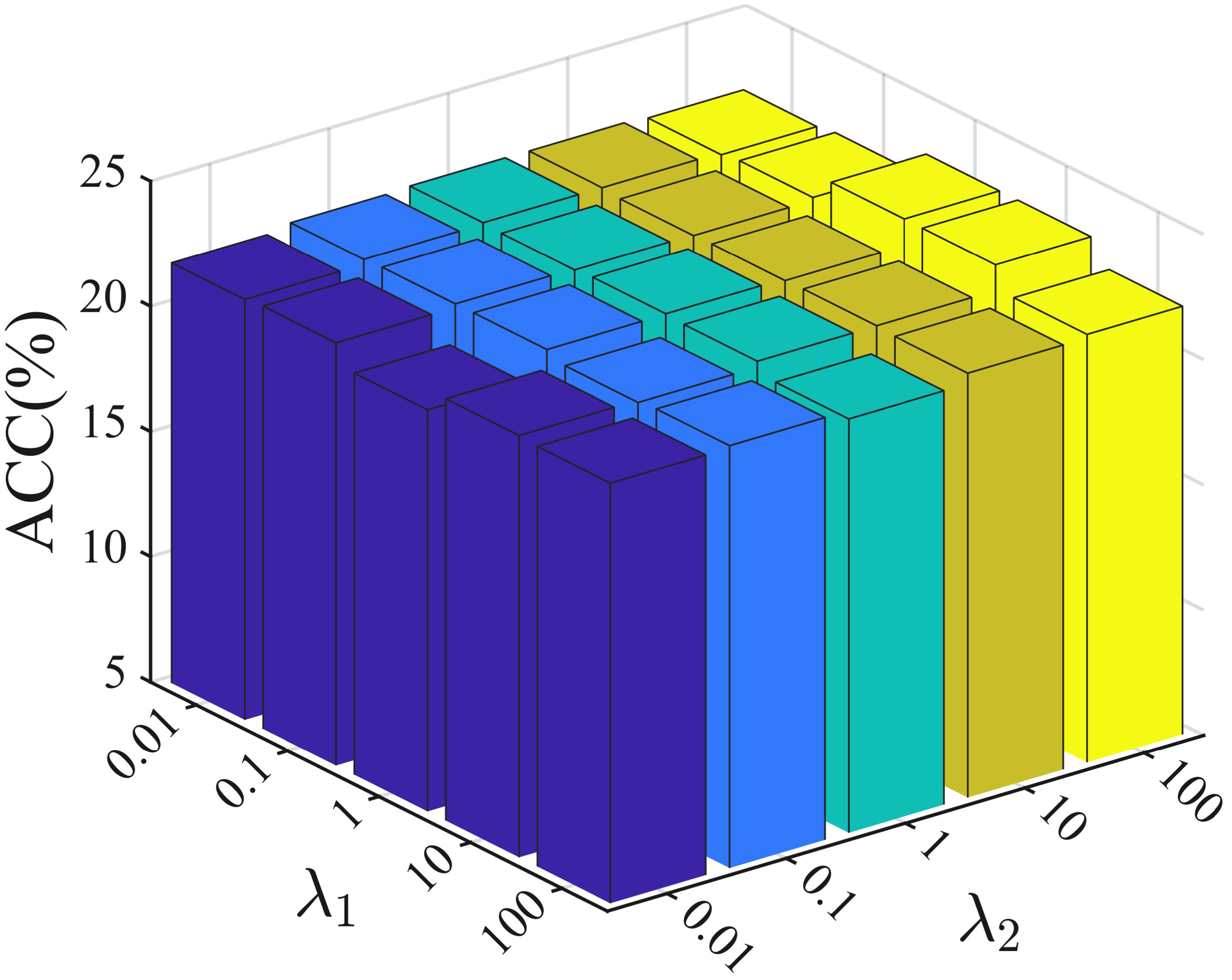}}
\subfigure[Caltech101-all]{
		\includegraphics[height=0.082\textheight,width=2.27cm]{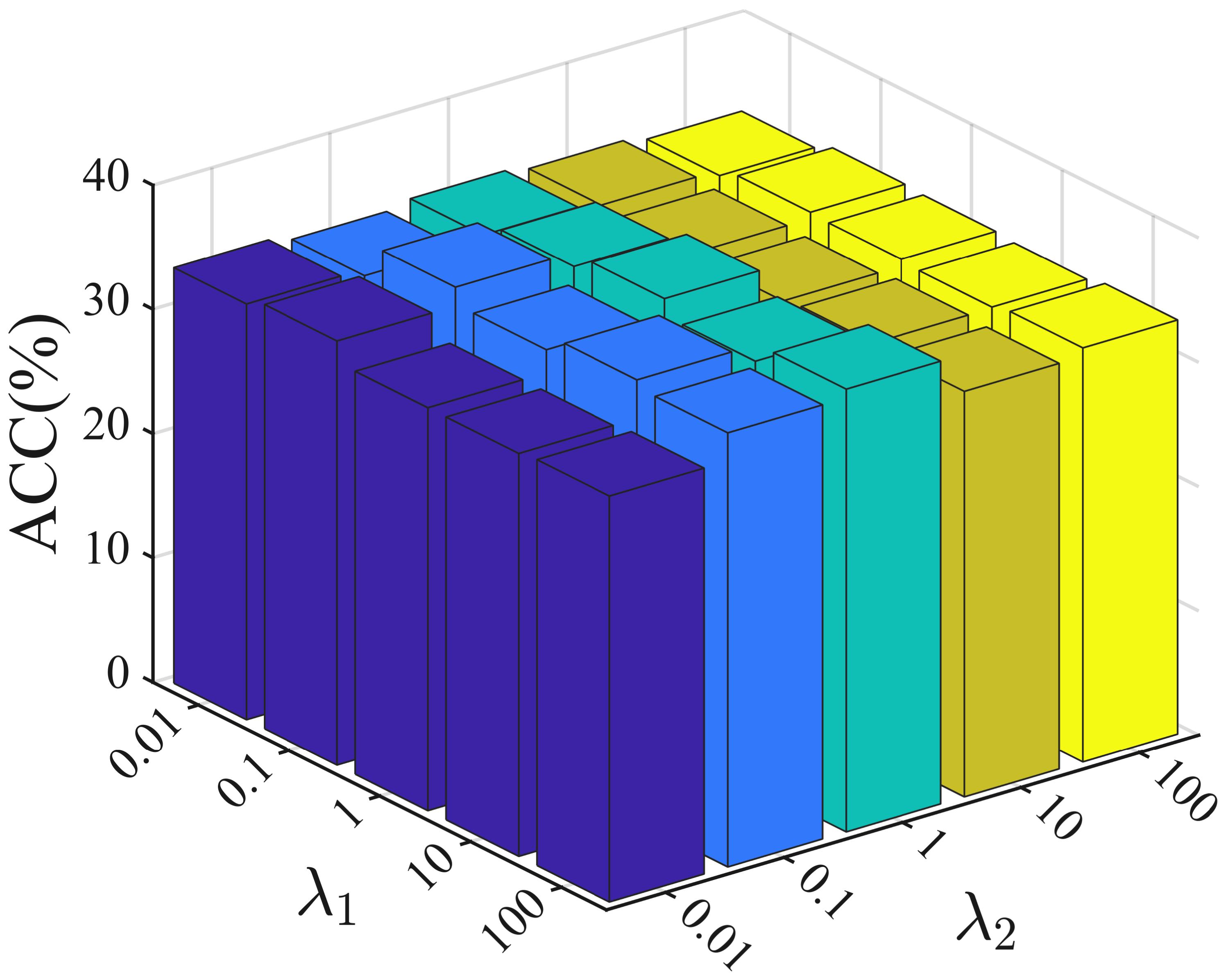}}
\subfigure[SUNRGBD]{
		\includegraphics[height=0.082\textheight,width=2.27cm]{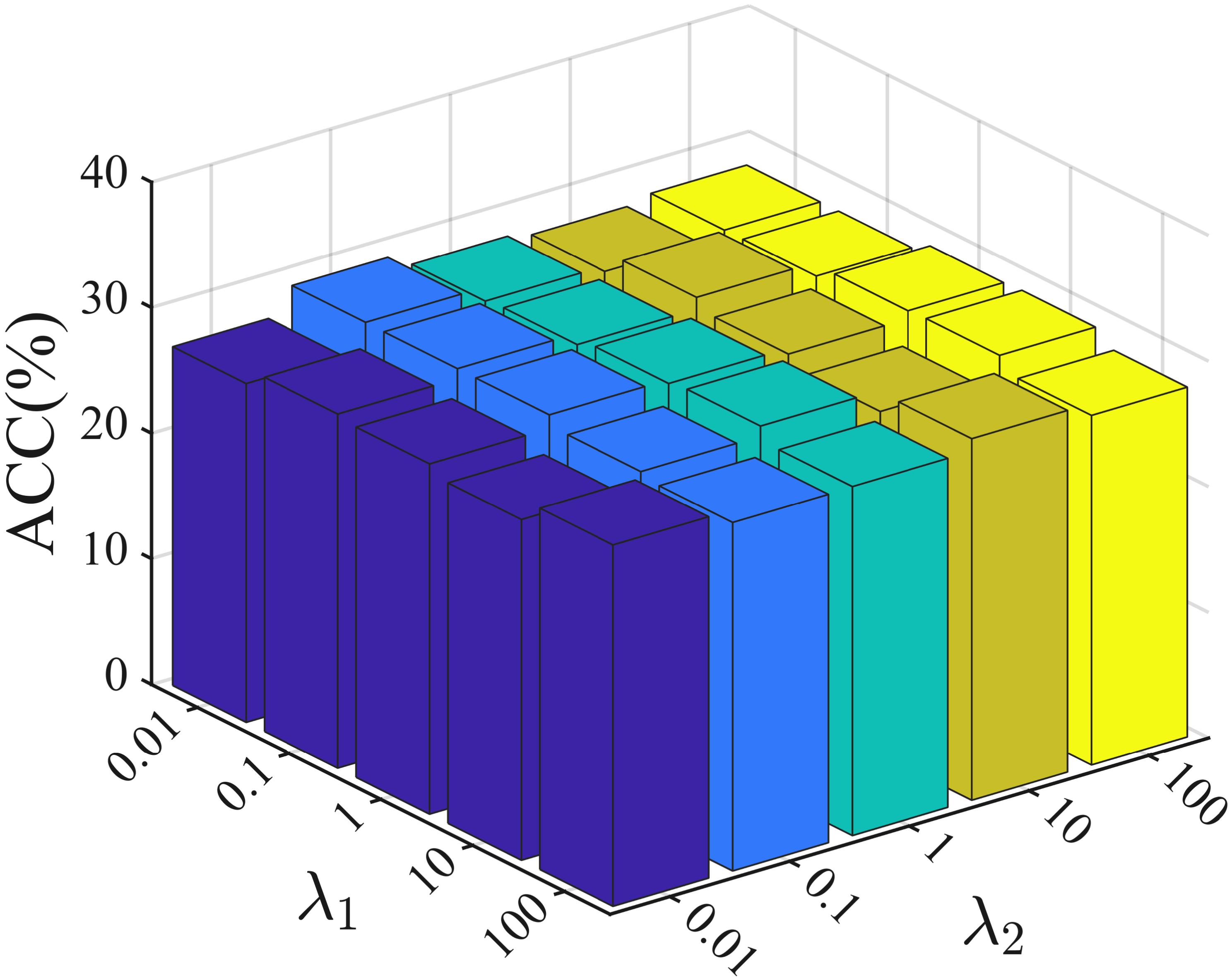}}
\subfigure[NUS-WIDE]{
		\includegraphics[height=0.082\textheight,width=2.27cm]{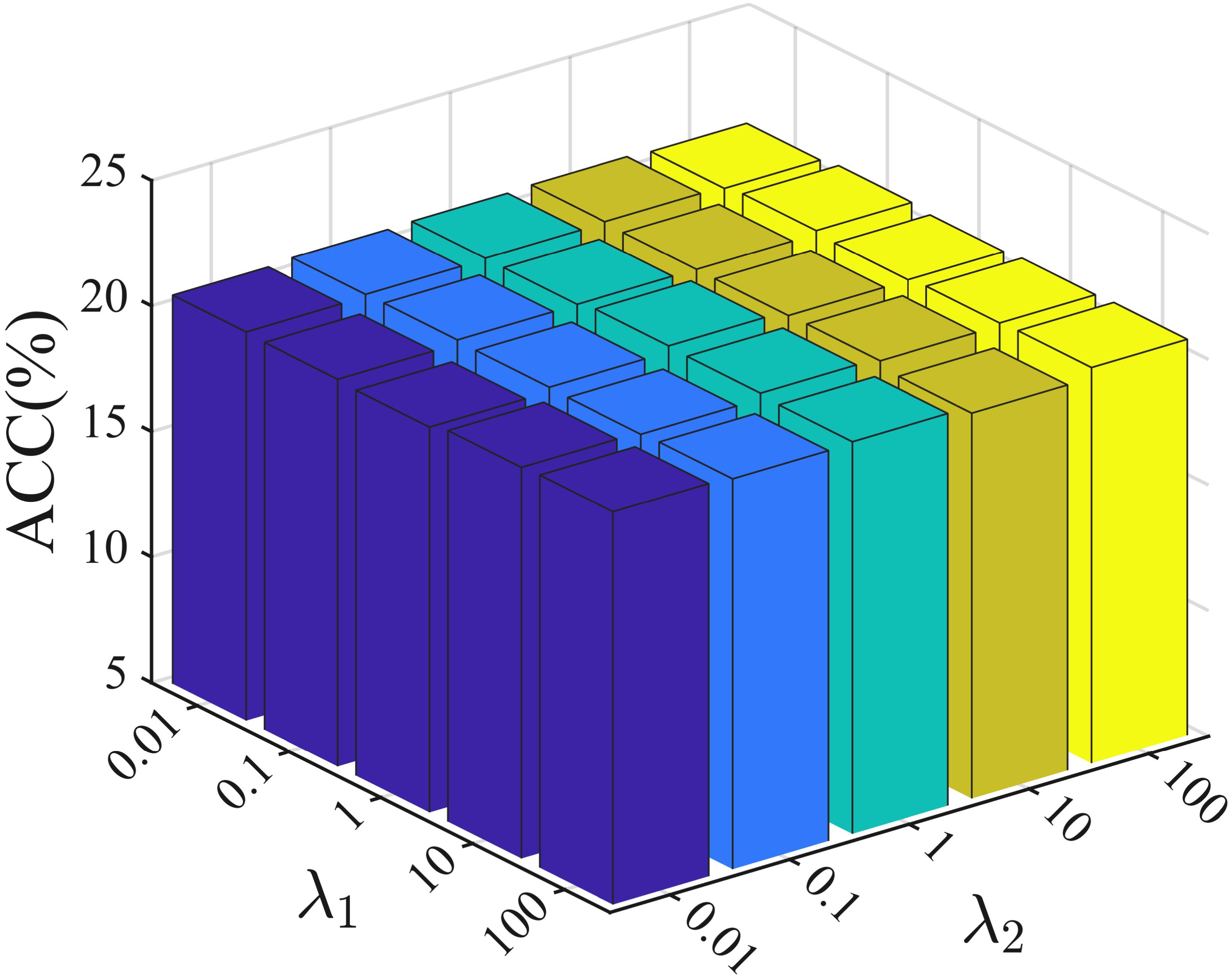}}
\subfigure[YouTubeFace]{
		\includegraphics[height=0.082\textheight,width=2.27cm]{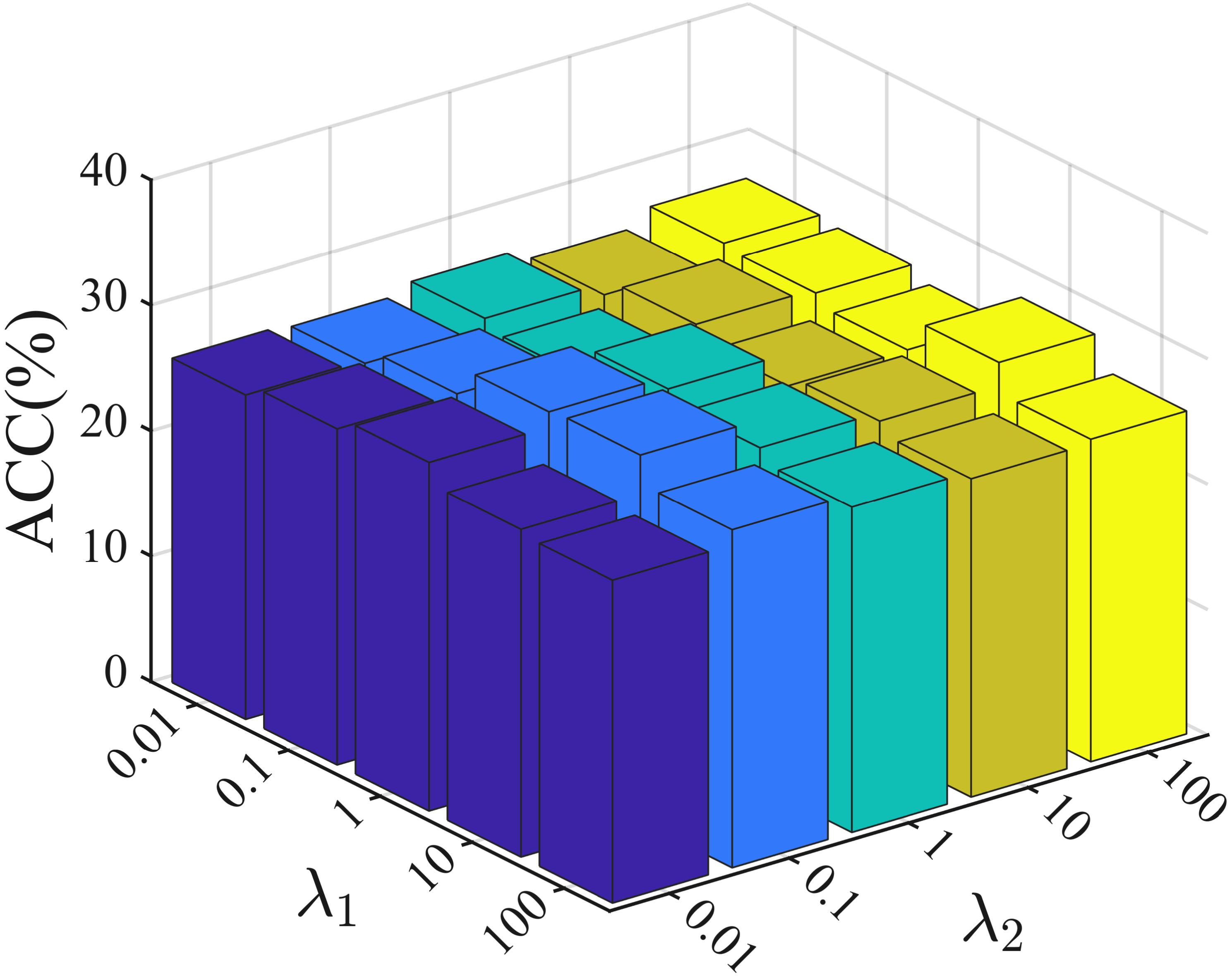}}

\caption{Clustering performance (w.r.t. ACC(\%)) with varying values of hyper-parameters ${\lambda }_{1}$ and ${\lambda }_{2}$ on seven datasets.  } 
\label{fig5}
\end{figure*}

\begin{figure*}[!t]
\centering
	\subfigcapskip=-3pt 
\subfigure[HW]{
		\includegraphics[height=0.082\textheight,width=2.27cm]{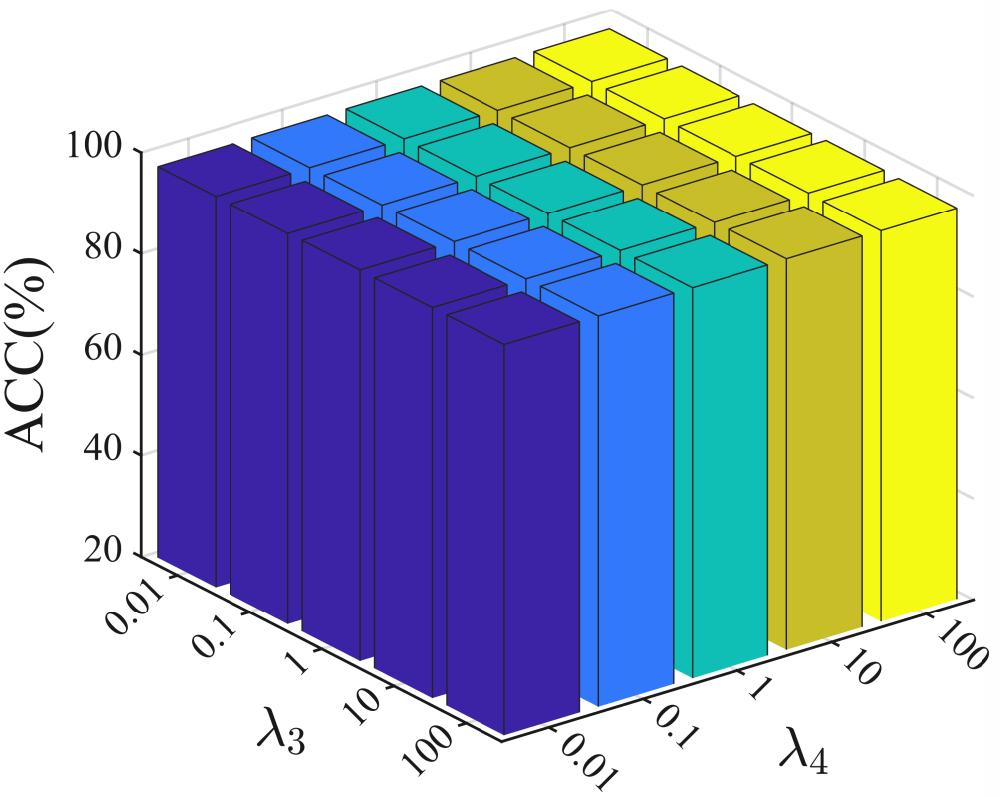}}
\subfigure[Caltech101-20]{
		\includegraphics[height=0.082\textheight,width=2.27cm]{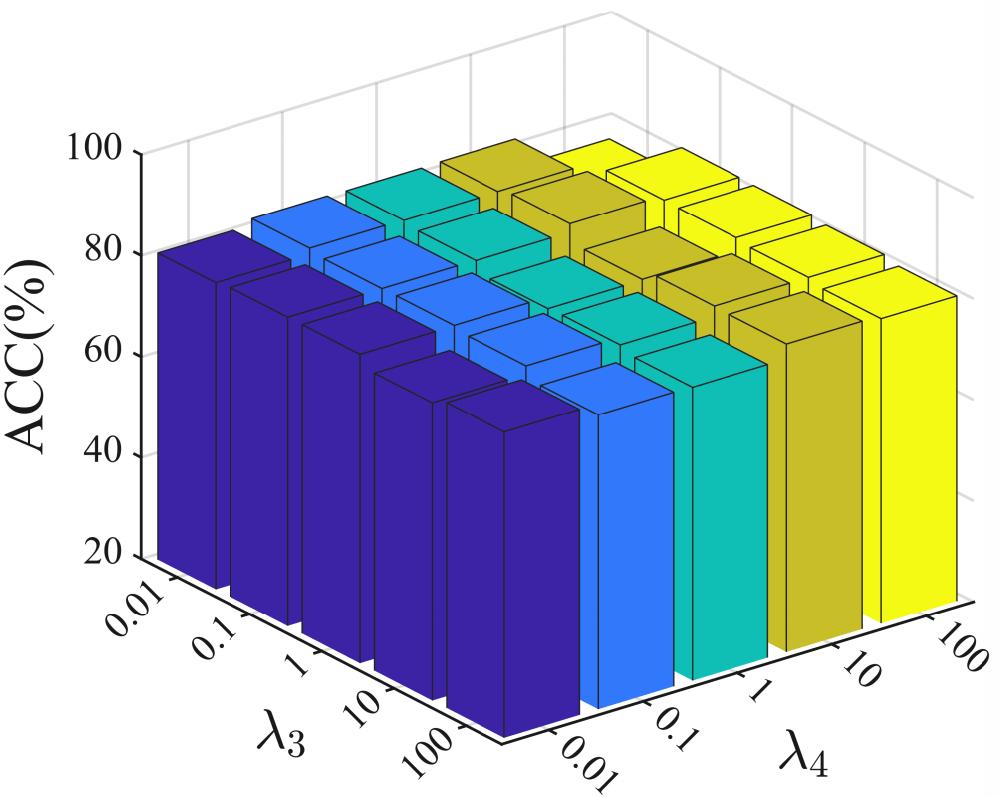}}
\subfigure[CCV]{
		\includegraphics[height=0.082\textheight,width=2.27cm]{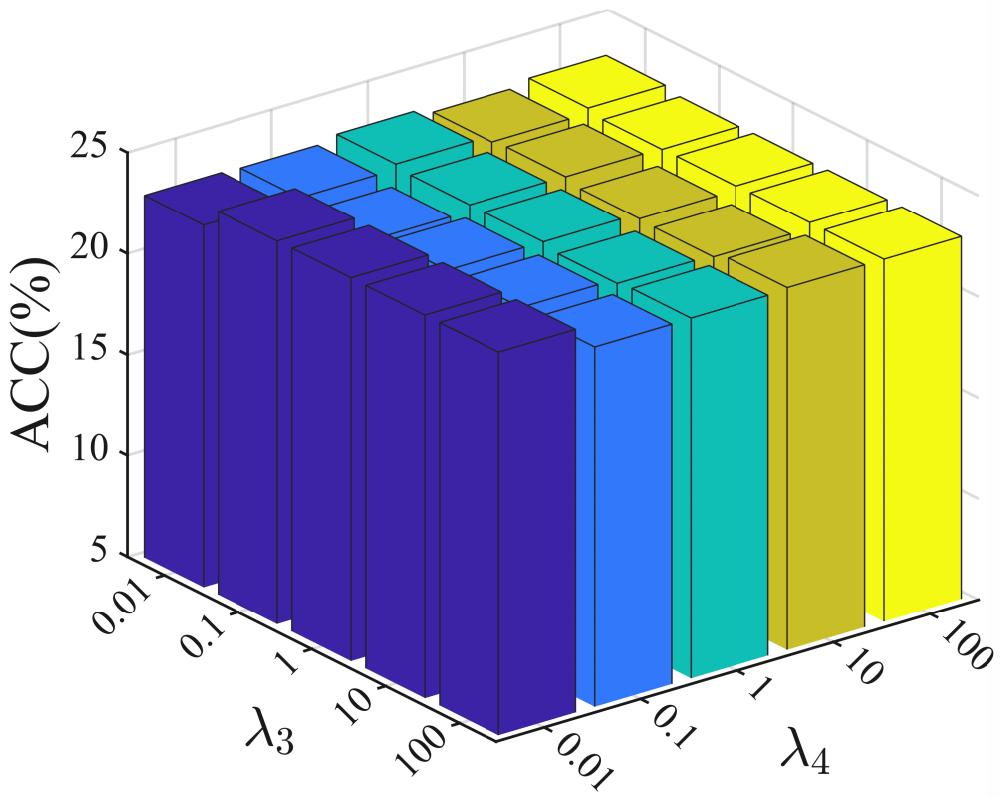}}
\subfigure[Caltech101-all]{
		\includegraphics[height=0.082\textheight,width=2.27cm]{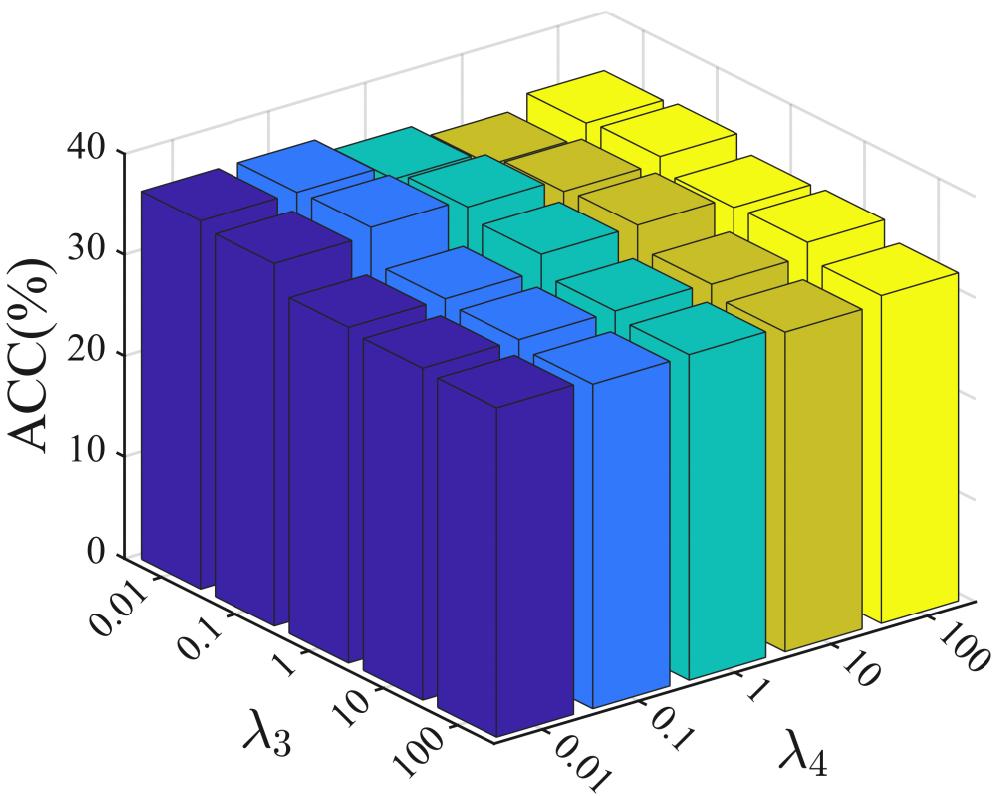}}
\subfigure[SUNRGBD]{
		\includegraphics[height=0.082\textheight,width=2.27cm]{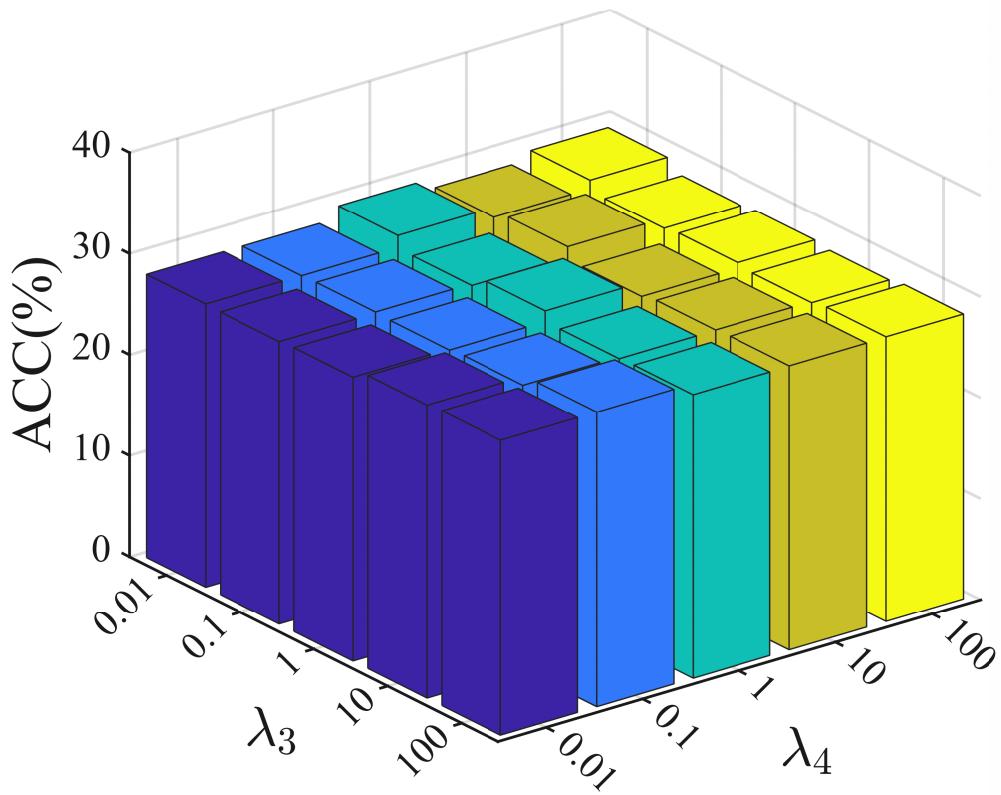}}
\subfigure[NUS-WIDE]{
		\includegraphics[height=0.082\textheight,width=2.27cm]{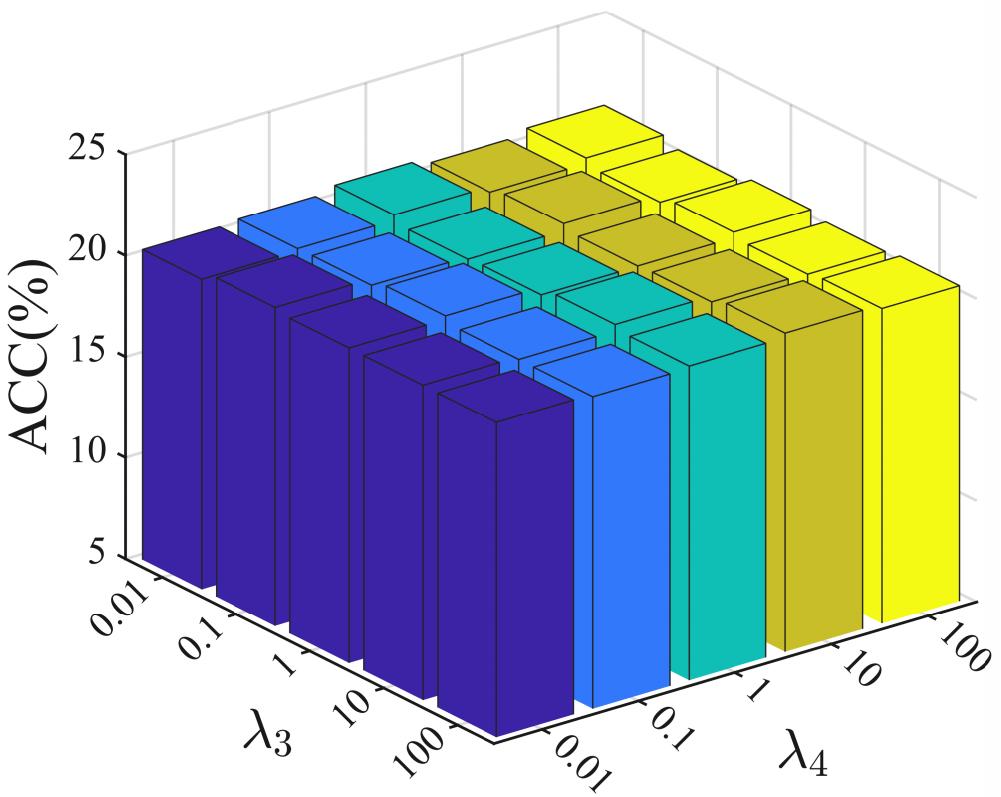}}
\subfigure[YouTubeFace]{
		\includegraphics[height=0.082\textheight,width=2.27cm]{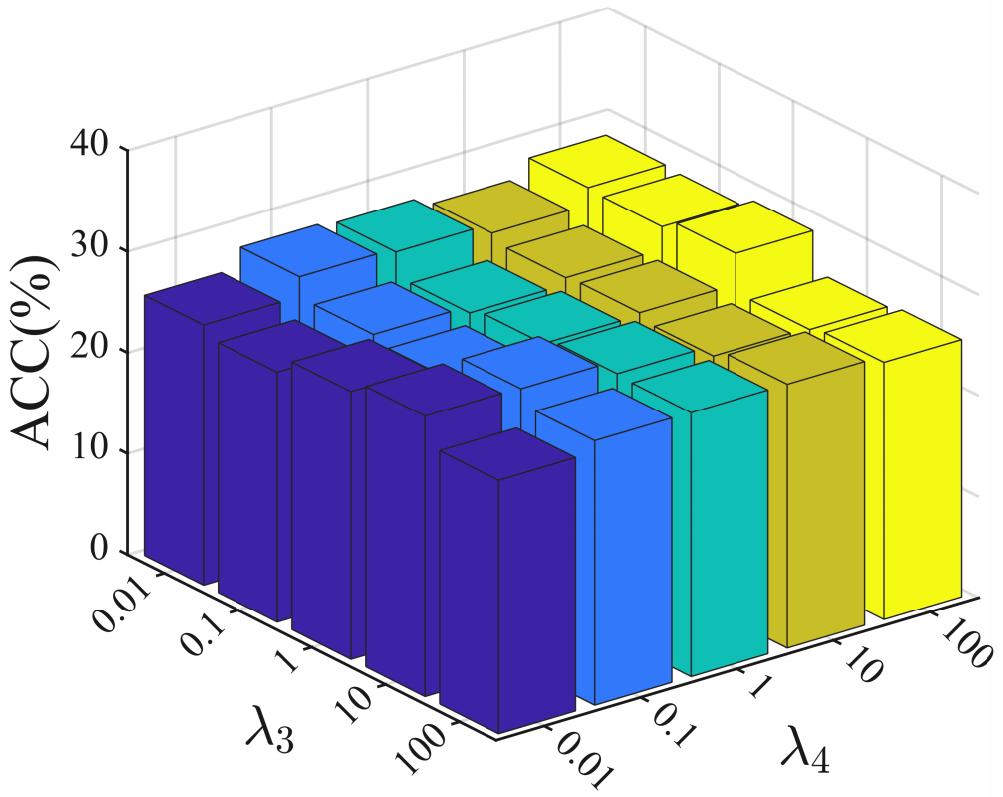}}

\caption{Clustering performance (w.r.t. ACC(\%)) with varying values of hyper-parameters ${\lambda }_{3}$ and ${\lambda }_{4}$ on seven datasets.}
\label{fig6}
\end{figure*}

\subsection{Parameter Sensitivity and Convergence Analysis}  \label{subsectionD}
\subsubsection{Parameter Sensitivity Analysis}  
As previously discussed, there are four trade-off parameters $\lambda_1,\lambda_2,\lambda_3,\lambda_4$ used to balance each loss component in our model. To analyze parameter sensitivity, the grid searching strategy is used to tune parameters within the range of $\{0.01, 0.1, 1.0, 10, 100\}$. Figs. \ref{fig5} and \ref{fig6} report the variation in clustering performance (w.r.t. ACC) with different parameter settings on all datasets. Concretely, through the controlling variables method, we begin by fixing parameters $\lambda_3$ and $\lambda_4$ to 1.0, while changing the values of $\lambda_1$ and $\lambda_2$. See Fig. \ref{fig5}. We can find that clustering performance remains relatively stable over a wide range of parameter combinations, indicating our E$^2$MVSC is insensitive to changes in $\lambda_1$ and $\lambda_2$ under most datasets. 

Similarly, we fix $\lambda_1=1.0$ and $\lambda_2=1.0$, adjust $\lambda_3$ and $\lambda_4$. From the results in Fig. \ref{fig6}, one could observe that the clustering result has not changed much as parameters vary. Indicating E$^2$MVSC still maintains robustness to the change of $\lambda_3$ and $\lambda_4$. Actually, based on parameter sensitivity analysis and the purpose of simplifying the parameter settings, we set $\lambda_1$, $\lambda_2$, $\lambda_3$, and $\lambda_4$ at 1.0 for all datasets in experiments, resulting in satisfying overall clustering performance.

 \begin{figure}[!t]
\centering
	\subfigcapskip=-3pt 
\subfigure[Hand Written]{
		\includegraphics[scale=0.2]{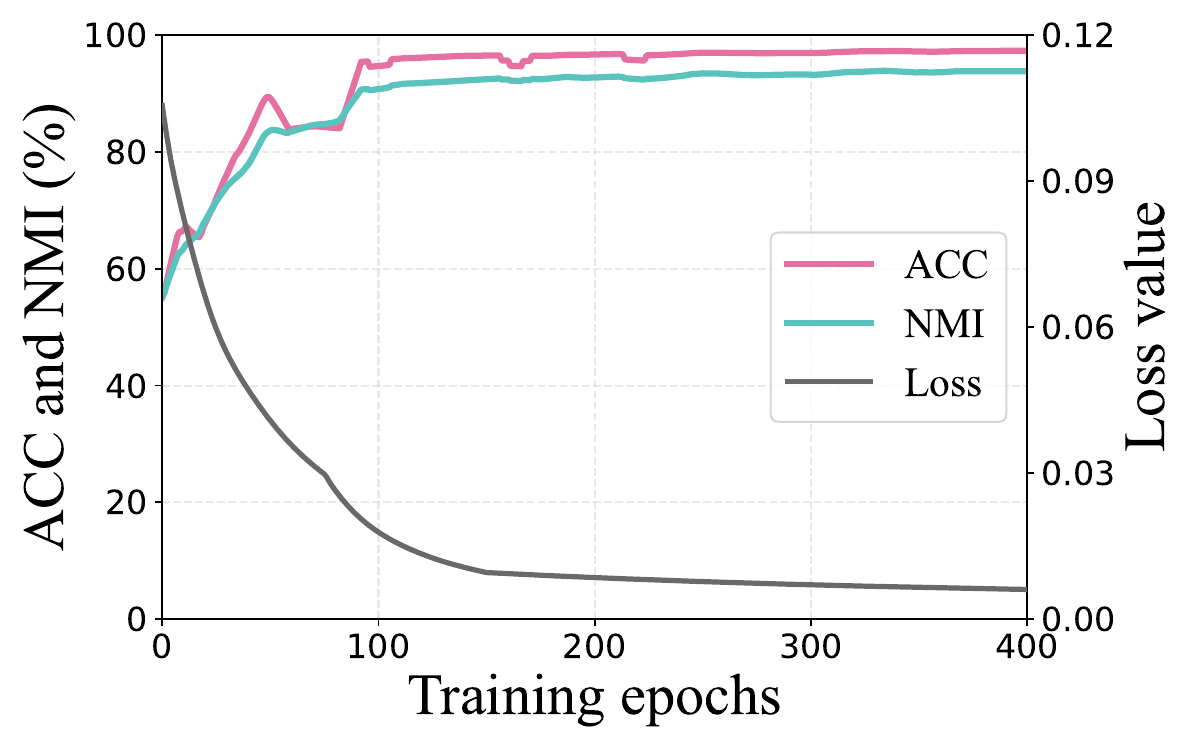}}
\subfigure[Caltech101-20]{
		\includegraphics[scale=0.2]{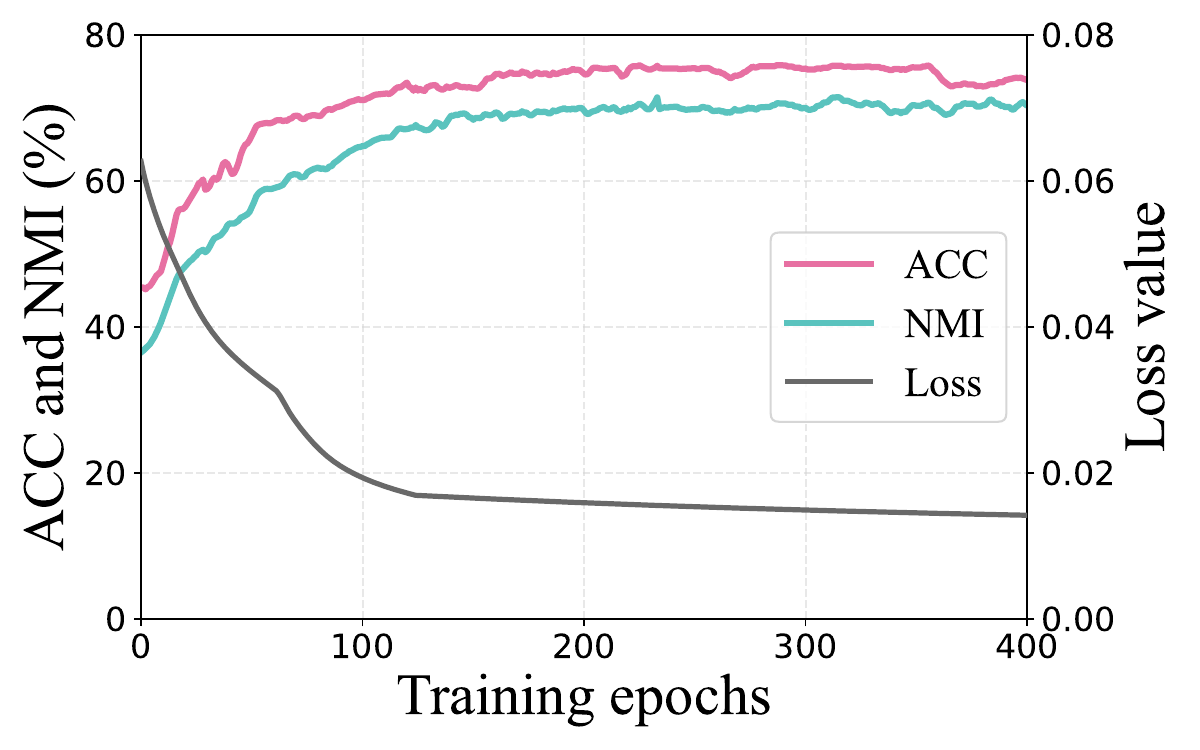}}

\caption{The objective value and clustering performance (ACC and NMI) vary with iterations on Hand Written and Caltech101-20 datasets.}
\label{fig7}
\end{figure}

\subsubsection{Convergence Analysis}  
The convergence curves, performed in Fig. \ref{fig7}, reflect the changes in total loss function values and clustering results of E$^2$MVSC with increasing epochs. Here, the Hand Written and Caltech101-20 datasets are used for analysis. It is evident that the loss function (see the gray line) drops rapidly in the initial few iterations, then decreases slowly along with increasing epochs until convergence. Typically, for these datasets, our model can converge in less than 300 epochs. Conversely, the clustering performance, measured by ACC and NMI, increases drastically and then tends to be stable as the objective function converges. Despite some slight shocks in the training process, overall, these results clearly verify that our proposed method possesses the desirable convergence property. Similar convergence results were obtained for other datasets but are not included here.

\section{Conclusion}
\label{Conclusion}
In this paper, we propose a novel deep multi-view subspace clustering method, termed E$^2$MVSC, for improving not only clustering effectiveness but also efficiency, especially for large-scale data cases. To the best of our knowledge, E$^2$MVSC is the first attempt to explicitly decouple multi-view information in latent space, including consistent, complementary, and superfluous information. This further promotes our model to retrieve a sufficient yet minimal unified feature representation following information bottleneck theory. Under the devised discriminative constraint, we also ensure within-cluster compactness and between-cluster separability of unified representation. Particularly, we study the challenge of high computational complexity in existing deep self-expression learning models. Distinct from such methods, we approach self-expression learning from a metric learning perspective, devising the Relation-Metric Net to achieve a big reduction in parameter scale and memory footprint. Extensive experiments on several multi-view datasets demonstrate the effectiveness and efficiency of our method.

Considering that practical applications often suffer from the absence of partial views across samples, we hope to deeply extend E$^2$MVSC to deal with incomplete multi-view clustering in future work.


\bibliographystyle{IEEEtran}
\bibliography{IEEEabrv,reference}

\begin{thebibliography}{10}
\providecommand{\url}[1]{#1}
\csname url@samestyle\endcsname
\providecommand{\newblock}{\relax}
\providecommand{\bibinfo}[2]{#2}
\providecommand{\BIBentrySTDinterwordspacing}{\spaceskip=0pt\relax}
\providecommand{\BIBentryALTinterwordstretchfactor}{4}
\providecommand{\BIBentryALTinterwordspacing}{\spaceskip=\fontdimen2\font plus
\BIBentryALTinterwordstretchfactor\fontdimen3\font minus \fontdimen4\font\relax}
\providecommand{\BIBforeignlanguage}[2]{{%
\expandafter\ifx\csname l@#1\endcsname\relax
\typeout{** WARNING: IEEEtran.bst: No hyphenation pattern has been}%
\typeout{** loaded for the language `#1'. Using the pattern for}%
\typeout{** the default language instead.}%
\else
\language=\csname l@#1\endcsname
\fi
#2}}
\providecommand{\BIBdecl}{\relax}
\BIBdecl

\bibitem{Wang2021FastPM}
S.~Wang, X.~Liu, X.~Zhu, P.~Zhang, Y.~Zhang, F.~Gao, and E.~Zhu, ``Fast parameter-free multi-view subspace clustering with consensus anchor guidance,'' \emph{IEEE Transactions on Image Processing}, vol.~31, pp. 556--568, 2022.

\bibitem{Peng2018StructuredAF}
X.~Peng, J.~Feng, S.~Xiao, W.-Y. Yau, J.~T. Zhou, and S.~Yang, ``Structured autoencoders for subspace clustering,'' \emph{IEEE Transactions on Image Processing}, vol.~27, pp. 5076--5086, 2018.

\bibitem{Ji2017DeepSC}
P.~Ji, T.~Zhang, H.~Li, M.~Salzmann, and I.~D. Reid, ``Deep subspace clustering networks,'' in \emph{Proceedings of International Conference on Neural Information Processing Systems}, 2017, pp. 24--33.

\bibitem{Xie2020JointDM}
Y.~Xie, B.~Lin, Y.~Qu, C.~Li, W.~Zhang, L.~Ma, Y.~Wen, and D.~Tao, ``Joint deep multi-view learning for image clustering,'' \emph{IEEE Transactions on Knowledge and Data Engineering}, vol.~33, pp. 3594--3606, 2020.

\bibitem{Wang2023BiNuclearTS}
S.~Wang, Z.~Lin, Q.~Cao, Y.~Cen, and Y.~Chen, ``Bi-nuclear tensor schatten-p norm minimization for multi-view subspace clustering,'' \emph{IEEE Transactions on Image Processing}, vol.~32, pp. 4059--4072, 2023.

\bibitem{DBLP:journals/tip/ChenMPZP23}
J.~Chen, H.~Mao, D.~Peng, C.~Zhang, and X.~Peng, ``Multiview clustering by consensus spectral rotation fusion,'' \emph{IEEE Transactions on Image Processing}, vol.~32, pp. 5153--5166, 2023.

\bibitem{ZhangFHCXTX20}
C.~Zhang, H.~Fu, Q.~Hu, X.~Cao, Y.~Xie, D.~Tao, and D.~Xu, ``Generalized latent multi-view subspace clustering,'' \emph{IEEE Transactions on Pattern Analysis and Machine Intelligence}, vol.~42, no.~1, pp. 86--99, 2020.

\bibitem{DBLP:journals/isci/YuLLLS23}
X.~Yu, H.~Liu, Y.~Lin, N.~Liu, and S.~Sun, ``Sample-level weights learning for multi-view clustering on spectral rotation,'' \emph{Inf. Sci.}, vol. 619, pp. 38--51, 2023.

\bibitem{SunDL21}
S.~Sun, W.~Dong, and Q.~Liu, ``Multi-view representation learning with deep gaussian processes,'' \emph{IEEE Transactions on Pattern Analysis and Machine Intelligence}, vol.~43, no.~12, pp. 4453--4468, 2021.

\bibitem{Cao2015ConstrainedMV}
X.~Cao, C.~Zhang, C.~Zhou, H.~Fu, and H.~Foroosh, ``Constrained multi-view video face clustering,'' \emph{IEEE Transactions on Image Processing}, vol.~24, pp. 4381--4393, 2015.

\bibitem{Li2022HighOrderCP}
Z.~Li, C.~Tang, X.~Zheng, X.~Liu, W.~Zhang, and E.~Zhu, ``High-order correlation preserved incomplete multi-view subspace clustering,'' \emph{IEEE Transactions on Image Processing}, vol.~31, pp. 2067--2080, 2022.

\bibitem{Tang2022ConstrainedTR}
Y.~Tang, Y.~Xie, C.~Zhang, and W.~Zhang, ``Constrained tensor representation learning for multi-view semi-supervised subspace clustering,'' \emph{IEEE Transactions on Multimedia}, vol.~24, pp. 3920--3933, 2022.

\bibitem{Zhou2020DualSM}
T.~Zhou, C.~Zhang, X.~Peng, H.~Bhaskar, and J.~Yang, ``Dual shared-specific multiview subspace clustering,'' \emph{IEEE Transactions on Cybernetics}, vol.~50, pp. 3517--3530, 2020.

\bibitem{Gao2015MultiviewSC}
H.~Gao, F.~Nie, X.~Li, and H.~Huang, ``Multi-view subspace clustering,'' in \emph{Proceedings of IEEE/CVF International Conference on Computer Vision}, 2015, pp. 4238--4246.

\bibitem{LuoZZC18}
S.~Luo, C.~Zhang, W.~Zhang, and X.~Cao, ``Consistent and specific multi-view subspace clustering,'' in \emph{Proceedings of AAAI Conference on Artificial Intelligence}, 2018, pp. 3730--3737.

\bibitem{Zheng2019FeatureCM}
Q.~Zheng, J.~Zhu, Z.~Li, S.~Pang, and J.~Wang, ``Feature concatenation multi-view subspace clustering,'' \emph{Neurocomputing}, vol. 379, pp. 89--102, 2019.

\bibitem{Zhang2017LatentMS}
C.~Zhang, Q.~Hu, H.~Fu, P.~F. Zhu, and X.~Cao, ``Latent multi-view subspace clustering,'' in \emph{Proceedings of IEEE/CVF Conference on Computer Vision and Pattern Recognition}, 2017, pp. 4333--4341.

\bibitem{Chen2020MultiViewCI}
M.~Chen, L.~Huang, C.~Wang, and D.~Huang, ``Multi-view clustering in latent embedding space,'' in \emph{Proceedings of AAAI Conference on Artificial Intelligence}, 2020, pp. 3513--3520.

\bibitem{Abavisani2018DeepMS}
M.~Abavisani and V.~M. Patel, ``Deep multimodal subspace clustering networks,'' \emph{IEEE Journal of Selected Topics in Signal Processing}, vol.~12, pp. 1601--1614, 2018.

\bibitem{Li2019ReciprocalMS}
R.~Li, C.~Zhang, H.~Fu, X.~Peng, J.~T. Zhou, and Q.~Hu, ``Reciprocal multi-layer subspace learning for multi-view clustering,'' in \emph{Proceedings of IEEE/CVF International Conference on Computer Vision}, 2019, pp. 8171--8179.

\bibitem{Wang2021DeepMS}
Q.~Wang, J.~Cheng, Q.~Gao, G.~Zhao, and L.~Jiao, ``Deep multi-view subspace clustering with unified and discriminative learning,'' \emph{IEEE Transactions on Multimedia}, vol.~23, pp. 3483--3493, 2021.

\bibitem{LuLZ21}
R.~Lu, J.~Liu, and X.~Zuo, ``Attentive multi-view deep subspace clustering net,'' \emph{Neurocomputing}, vol. 435, pp. 186--196, 2021.

\bibitem{Wang2022SelfSupervisedIB}
S.~Wang, C.~Li, Y.~Li, Y.~Yuan, and G.~Wang, ``Self-supervised information bottleneck for deep multi-view subspace clustering,'' \emph{IEEE Transactions on Image Processing}, vol.~32, pp. 1555--1567, 2022.

\bibitem{DBLP:journals/tcyb/LiLZLF22}
K.~Li, H.~Liu, Y.~Zhang, K.~Li, and Y.~Fu, ``Self-guided deep multiview subspace clustering via consensus affinity regularization,'' \emph{IEEE Transactions on Cybernetics}, vol.~52, no.~12, pp. 12\,734--12\,744, 2022.

\bibitem{KangZZSHX20}
Z.~Kang, W.~Zhou, Z.~Zhao, J.~Shao, M.~Han, and Z.~Xu, ``Large-scale multi-view subspace clustering in linear time,'' in \emph{Proceedings of AAAI Conference on Artificial Intelligence}, 2020, pp. 4412--4419.

\bibitem{Sun2021ScalableMS}
M.~Sun, P.~Zhang, S.~Wang, S.~Zhou, W.~Tu, X.~Liu, E.~Zhu, and C.~Wang, ``Scalable multi-view subspace clustering with unified anchors,'' in \emph{Proceedings of ACM International Conference on Multimedia}, 2021, pp. 3528--3536.

\bibitem{Li2020MultiviewCA}
X.~Li, H.~Zhang, R.~Wang, and F.~Nie, ``Multiview clustering: A scalable and parameter-free bipartite graph fusion method,'' \emph{IEEE Transactions on Pattern Analysis and Machine Intelligence}, vol.~44, pp. 330--344, 2020.

\bibitem{Zhang2019AE2NetsAI}
C.~Zhang, Y.~Liu, and H.~Fu, ``Ae2-nets: Autoencoder in autoencoder networks,'' in \emph{Proceedings of IEEE/CVF Conference on Computer Vision and Pattern Recognition}, 2019, pp. 2572--2580.

\bibitem{Zheng2022ComprehensiveMR}
Q.~Zheng, J.~Zhu, Z.~Li, Z.~Tian, and C.~Li, ``Comprehensive multi-view representation learning,'' \emph{Inf. Fusion}, vol.~89, pp. 198--209, 2022.

\bibitem{Wang2020iCmSCIC}
Q.~Wang, H.~Lian, G.~Sun, Q.~Gao, and L.~Jiao, ``icmsc: Incomplete cross-modal subspace clustering,'' \emph{IEEE Transactions on Image Processing}, vol.~30, pp. 305--317, 2020.

\bibitem{XuT0P0022}
J.~Xu, H.~Tang, Y.~Ren, L.~Peng, X.~Zhu, and L.~He, ``Multi-level feature learning for contrastive multi-view clustering,'' in \emph{Proceedings of IEEE/CVF Conference on Computer Vision and Pattern Recognition}, 2022, pp. 16\,030--16\,039.

\bibitem{Lin2023DualCP}
Y.~Lin, Y.~Gou, X.~Liu, J.~Bai, J.~Lv, and X.~Peng, ``Dual contrastive prediction for incomplete multi-view representation learning,'' \emph{IEEE Transactions on Pattern Analysis and Machine Intelligence}, vol.~45, pp. 4447--4461, 2023.

\bibitem{Cai2022UnsupervisedDD}
J.~Cai, W.~Guo, and J.~Fan, ``Unsupervised deep discriminant analysis based clustering,'' \emph{ArXiv}, vol. abs/2206.04686, 2022.

\bibitem{Fu2023MultiViewCF}
L.~Fu, L.~Zhang, T.~Wang, C.~Chen, C.~Zhang, and Z.~Zheng, ``Multi-view clustering from the perspective of mutual information,'' \emph{ArXiv}, vol. abs/2302.08743, 2023.

\bibitem{DBLP:conf/iclr/Federici0FKA20}
M.~Federici, A.~Dutta, P.~Forr{\'{e}}, N.~Kushman, and Z.~Akata, ``Learning robust representations via multi-view information bottleneck,'' in \emph{Proceedings of ICLR}, 2020.

\bibitem{Yu2020LearningDA}
Y.~Yu, K.~H.~R. Chan, C.~You, C.~Song, and Y.~Ma, ``Learning diverse and discriminative representations via the principle of maximal coding rate reduction,'' in \emph{Proceedings of International Conference on Neural Information Processing Systems}, 2020.

\bibitem{Huang2019MultiviewSC}
Z.~Huang, J.~T. Zhou, X.~Peng, C.~Zhang, H.~Zhu, and J.~Lv, ``Multi-view spectral clustering network,'' in \emph{Proceedings of International Joint Conference on Artificial Intelligence}, 2019, pp. 2563--2569.

\bibitem{Xu2023SelfSupervisedDF}
J.~Xu, Y.~Ren, H.~Tang, Z.~Yang, L.~Pan, Y.~Yang, and X.~Pu, ``Self-supervised discriminative feature learning for deep multi-view clustering,'' \emph{IEEE Transactions on Knowledge and Data Engineering}, vol.~35, no.~7, pp. 7470--7482, 2023.

\bibitem{Liu2021OnepassMC}
J.~Liu, X.~Liu, Y.~Yang, L.~Liu, S.~Wang, W.~Liang, and J.~Shi, ``One-pass multi-view clustering for large-scale data,'' in \emph{Proceedings of IEEE/CVF International Conference on Computer Vision}, 2021, pp. 12\,324--12\,333.

\bibitem{Wan2023AutoweightedMC}
X.~Wan, X.~Liu, J.~Liu, S.~Wang, Y.~Wen, W.~Liang, E.~Zhu, Z.~Liu, and L.~Zhou, ``Auto-weighted multi-view clustering for large-scale data,'' in \emph{Proceedings of AAAI Conference on Artificial Intelligence}, 2023, pp. 10\,078--10\,086.

\bibitem{Zhang2023CenterCG}
X.~Zhang, Z.~Ren, and C.~Yang, ``Center consistency guided multi-view embedding anchor learning for large-scale graph clustering,'' \emph{Knowl. Based Syst.}, vol. 260, p. 110162, 2023.

\bibitem{Liu2022EfficientOM}
S.~Liu, S.~Wang, P.~Zhang, K.~Xu, X.~Liu, C.~Zhang, and F.~Gao, ``Efficient one-pass multi-view subspace clustering with consensus anchors,'' in \emph{Proceedings of AAAI Conference on Artificial Intelligence}, 2022.

\bibitem{Busch2020LearningSM}
J.~Busch, E.~Faerman, M.~Schubert, and T.~Seidl, ``Learning self-expression metrics for scalable and inductive subspace clustering,'' \emph{ArXiv}, vol. abs/2009.12875, 2020.

\bibitem{ZhangYVL21}
S.~Zhang, C.~You, R.~Vidal, and C.~Li, ``Learning a self-expressive network for subspace clustering,'' in \emph{Proceedings of IEEE/CVF Conference on Computer Vision and Pattern Recognition}, 2021, pp. 12\,393--12\,403.

\bibitem{Mao2021DeepMI}
Y.~Mao, X.~Yan, Q.~Guo, and Y.~Ye, ``Deep mutual information maximin for cross-modal clustering,'' in \emph{Proceedings of AAAI Conference on Artificial Intelligence}, 2021.

\bibitem{Wan2021MultiViewIR}
Z.~Wan, C.~Zhang, P.~F. Zhu, and Q.~Hu, ``Multi-view information-bottleneck representation learning,'' in \emph{Proceedings of AAAI Conference on Artificial Intelligence}, 2021.

\bibitem{Wang_2023_CVPR}
R.~Wang, H.~Sun, Y.~Ma, X.~Xi, and Y.~Yin, ``Metaviewer: Towards a unified multi-view representation,'' in \emph{Proceedings of IEEE/CVF Conference on Computer Vision and Pattern Recognition}, 2023, pp. 11\,590--11\,599.

\bibitem{Yan2023MultiviewSC}
W.~biao Yan, J.~Zhu, Y.~Zhou, Y.~Wang, and Q.~Zheng, ``Multi-view semantic consistency based information bottleneck for clustering,'' \emph{ArXiv}, vol. abs/2303.00002, 2023.

\bibitem{Ren2022DeepCA}
Y.~Ren, J.~Pu, Z.~Yang, J.~Xu, G.~Li, X.~Pu, P.~S. Yu, and L.~He, ``Deep clustering: A comprehensive survey,'' \emph{ArXiv}, vol. abs/2210.04142, 2022.

\bibitem{Zhou2023SemanticallyCM}
Y.~Zhou, Q.~Zheng, S.~Bai, and J.~Zhu, ``Semantically consistent multi-view representation learning,'' \emph{Knowl. Based Syst.}, vol. 278, p. 110899, 2023.

\bibitem{Barber2003TheIA}
D.~Barber and F.~V. Agakov, ``The im algorithm: a variational approach to information maximization,'' in \emph{Proceedings of International Conference on Neural Information Processing Systems}, 2003.

\bibitem{shapiro2003monte}
A.~Shapiro, ``Monte carlo sampling methods,'' \emph{Handbooks in operations research and management science}, vol.~10, pp. 353--425, 2003.

\bibitem{Lu2012RobustAE}
C.~Lu, H.~Min, Z.~Zhao, L.~Zhu, D.~shuang Huang, and S.~Yan, ``Robust and efficient subspace segmentation via least squares regression,'' in \emph{Proceedings of European Conference on Computer Vision}, 2012.

\end{thebibliography}

\vfill

\end{document}